%% file: main.tex
\documentclass[10pt,twocolumn,letterpaper]{article}

\usepackage[pagenumbers]{cvpr} % To force page numbers, e.g. for an arXiv version

\input{preamble}

\definecolor{cvprblue}{rgb}{0.21,0.49,0.74}
\usepackage[pagebackref,breaklinks,colorlinks,allcolors=cvprblue]{hyperref}

\def\paperID{} % *** Enter the Paper ID here
\def\confName{CVPR}
\def\confYear{2026}

\title{ \raisebox{-0.3\height}{\includegraphics[height=1.5em]{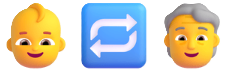}} Face Time Traveller : Travel Through Ages Without Losing Identity}

\author{
Purbayan Kar$^1$ \quad Ayush Ghadiya$^1$ \quad Vishal Chudasama$^1$ \quad Pankaj Wasnik$^1$ \quad C.V. Jawahar$^2$ \\ 
$^1$Sony Research India \quad $^2$IIIT Hyderabad\\ 
{\tt\small \{purbayan.kar, ayush.ghadiya, vishal.chudasama1, pankaj.wasnik\}@sony.com, jawahar@iiit.ac.in} 
}

\begin{document}
\maketitle

\input{sec/0_abstract}

\input{sec/1_intro}

{
    \small
    \bibliographystyle{ieeenat_fullname}
    \bibliography{main}
}

\input{sec/X_suppl}

\end{document}

%% file: preamble.tex
\usepackage{tabularx}
\usepackage{algorithm}
\usepackage{algpseudocode}
\usepackage[table]{xcolor}
\usepackage{soul}
\usepackage{adjustbox}
\usepackage{graphicx}
\usepackage{multirow}
\usepackage{pifont}
\usepackage{graphicx}
\usepackage[table]{xcolor}
\definecolor{mintbg}{rgb}{.63,.79,.95}
\colorlet{lightmintbg}{mintbg!30}
\definecolor{celadon}{rgb}{0.68, 0.87, 0.68}

\newcommand{\StartSuppleTOC}{
  \begingroup
  \let\oldsection\section
  \let\oldsubsection\subsection
  \let\oldsubsubsection\subsubsection
  \makeatletter
  \def\section{\addtocontents{suppletoc}{\protect\contentsline {section}{\protect\numberline {\thesection}}{\thepage}{section*}}\oldsection}
  \def\subsection{\addtocontents{suppletoc}{\protect\contentsline {subsection}{\protect\numberline {\thesubsection}}{\thepage}{subsection*}}\oldsubsection}
  \def\subsubsection{\addtocontents{suppletoc}{\protect\contentsline {subsubsection}{\protect\numberline {\thesubsubsection}}{\thepage}{subsubsection*}}\oldsubsubsection}
  \makeatother
}
\newcommand{\PrintSuppleTOC}{
  \begingroup
    \makeatletter
    \@starttoc{suppletoc}
    \makeatother
  \endgroup
}

%% file: sec/0_abstract.tex
\begin{abstract}
Face aging, an ill-posed problem shaped by environmental and genetic factors, is vital in entertainment, forensics, and digital archiving, where realistic age transformations must preserve both identity and visual realism. However, existing works relying on numerical age representations overlook the interplay of biological and contextual cues. Despite progress in recent face aging models, they struggle with identity preservation in wide age transformations, also static attention and optimization-heavy inversion in diffusion limit adaptability, fine-grained control and background consistency. To address these challenges, we propose Face Time Traveller (FaceTT), a diffusion-based framework that achieves high-fidelity, identity-consistent age transformation. Here, we introduce a Face-Attribute-Aware Prompt Refinement strategy that encodes intrinsic (biological) and extrinsic (environmental) aging cues for context-aware conditioning. A tuning-free Angular Inversion method is proposed that efficiently maps real faces into the diffusion latent space for fast and accurate reconstruction. Moreover, an Adaptive Attention Control mechanism is introduced that dynamically balances cross-attention for semantic aging cues and self-attention for structural and identity preservation. Extensive experiments on benchmark datasets and in-the-wild testset demonstrate that FaceTT achieves superior identity retention, background preservation and aging realism over state-of-the-art (SOTA) methods.
% show that FaceTT significantly enhances identity retention, background stability, and aging realism, outperforming state-of-the-art methods both quantitatively and qualitatively.
\end{abstract}

%% file: sec/1_intro.tex
\vspace{-10pt}
\section{Introduction} \label{sec:intro}
\begin{figure}[t]
    \centering
    \includegraphics[width=0.47\textwidth, height=8.5cm]{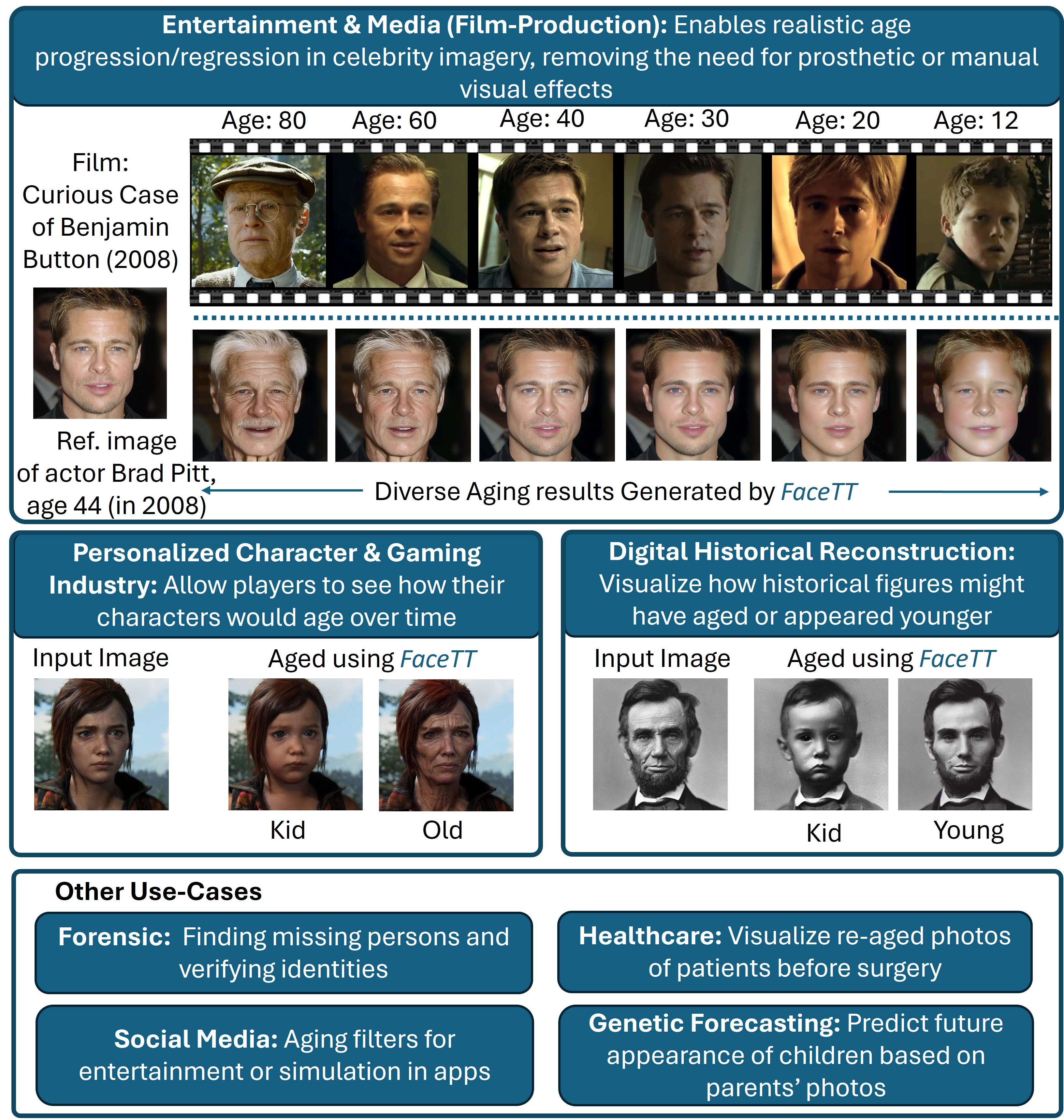}
    \vspace{-0.7em}
    % \caption{Diverse real-world use-cases of face re-aging. 
    \caption{Illustration of real-world use cases of face aging. \textbf{Top:} The Benjamin Button example from \textit{The Curious Case of Benjamin Button (2008)}, where Brad Pitt was digitally aged using a complex hybrid VFX pipeline \cite{danks2020not}. Modern face aging models can achieve similar visual realism at significantly lower time and cost without prosthetics or manual VFX—while preserving the actor’s identity across different lifespans. \textbf{Middle (left):} Application in personalized character and gaming environments. \textbf{Middle (right):} Digital historical reconstruction. \textbf{Bottom:} Additional use cases.
    % include forensic analysis, healthcare visualization, social-media filters, and genetic forecasting.
    % Right: Additional applications include forensics (e.g., missing person identification), healthcare simulations, social media aging filters, and genetic forecasting based on parental photos. These examples underline the versatility of our approach across creative, forensic, and predictive domains.
    }
    \label{fig:reaging_application}
    \vspace{-1em}
\end{figure}

\begin{figure}[t]
    \centering
    \includegraphics[width=0.47\textwidth]{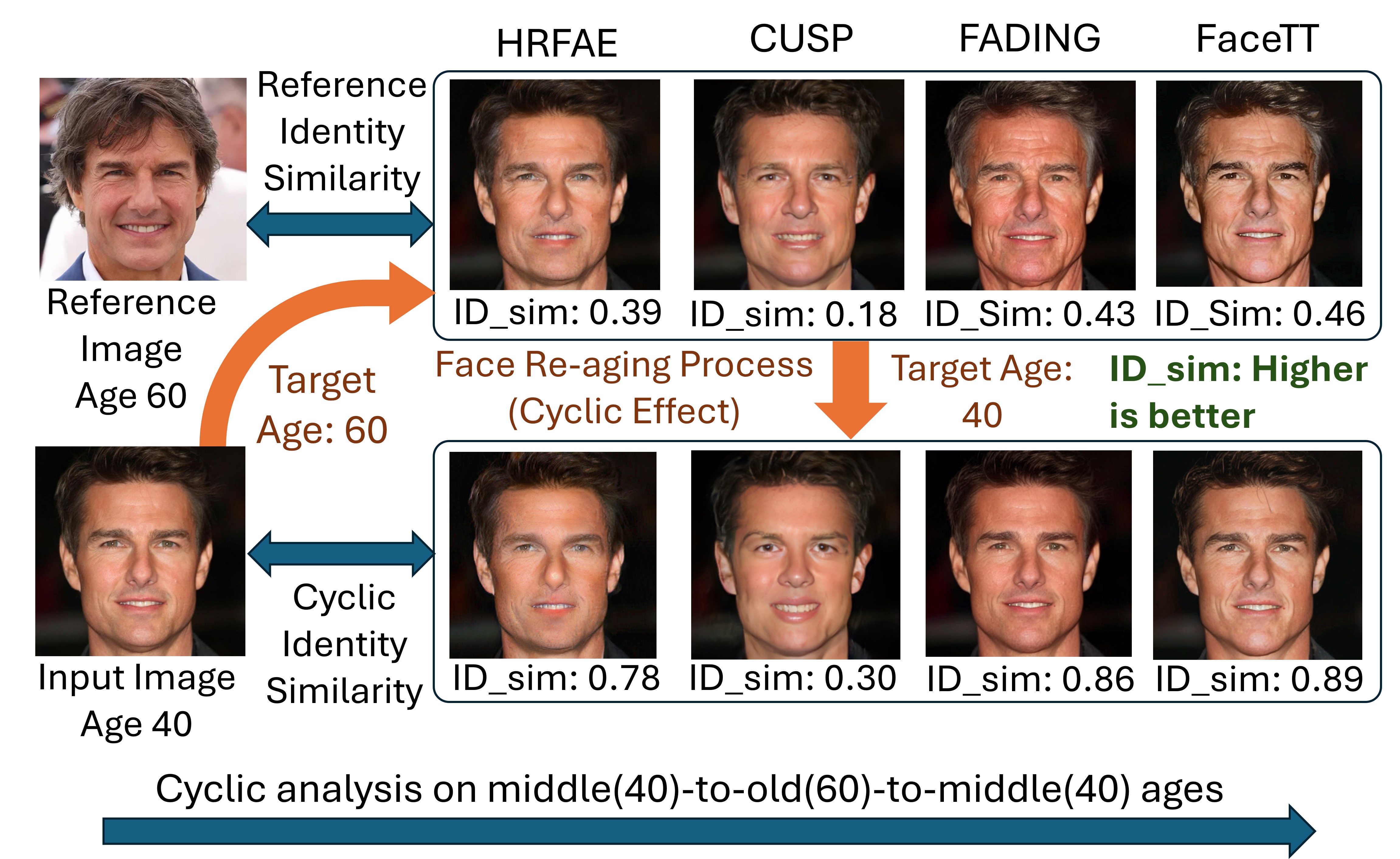}
    \vspace{-0.5em}
    % \caption{Diverse real-world use-cases of face re-aging. 
    \caption{Illustration of cyclic identity similarity protocol. Input image (age 40) is first re-aged to the target age (60) and then reverted to its original age (40). Top row depicts the reference-based identity similarity between the re-aged and reference image, while bottom row shows the cyclic identity similarity between the input and reconstructed faces.
    % Cyclic-consistency results across celebrity images shows strong identity preservation under extreme age transformations from middle-old-middle age group (40-60-40).
    }
    \label{fig:intro-cylic}
    \vspace{-1.5em}
\end{figure}

Face aging, the process of simulating age progression or regression in face images, has numerous real-world applications, from digital entertainment~\cite{loock2021realist, FRAN, hrfae} and medical analysis to forensic investigations~\cite{forensics}. As shown in Figure~\ref{fig:reaging_application}, its use extends across entertainment media, gaming, and cultural preservation. In film and advertising, productions such as \textit{The Irishman} and public campaigns like the David Beckham malaria awareness project demonstrate the demand for realistic age transformation \cite{FRAN}. Traditional approaches often rely on costly and labor-intensive VFX pipelines. In gaming, face aging allows characters to evolve over time, and in heritage and archival work, it supports visualizing historical figures across different life stages. Recent research advances~\cite{FADING, hrfae, cusp, FRAN} now make high-quality face aging more accessible, reducing both time and computational cost compared to conventional workflows.
% As illustrated in Figure~\ref{fig:reaging_application}, the usage of face aging has grown significantly across creative media, gaming, and cultural preservation. In the media sector, high-budget productions like \textit{The Irishman}, which digitally de-aged Robert De Niro, and public health campaigns that aged David Beckham to raise malaria awareness highlight the demand for realistic face aging. Traditionally, such transformations have relied on costly, labor-intensive pipelines like VFX-CGI. The gaming industry leverages face aging to create dynamic characters that evolve over time, while digital reconstruction projects visualize historical figures across different life stages for education and archival purposes. Although traditional approaches remain time-consuming and expensive, recent research methods~\cite{FADING, hrfae, cusp, FRAN} now enable high-quality aging with substantially lower time and computational cost.

Despite promising advances, producing convincing aging results remains a complex and ill-posed problem. It requires a careful balance between altering age-relevant features and preserving age agnostic ones, such as facial identity and expression. 
The challenge intensifies when dealing with significant age changes, diverse facial characteristics, and varying image conditions such as occlusions or complex backgrounds. It also involves understanding the aging process influenced by both internal factors—like genetics, hormones, and metabolism—and external ones such as UV exposure, lifestyle, and environmental effects. While intrinsic and extrinsic factors play a fundamental role in the biological aging process, this work considers their combined visual manifestations by addressing structural diversity, illumination variations, and background complexity, all while rigorously preserving the facial identity.

Recent advancements using generative adversarial networks have shown promise in addressing the face aging problem~\cite{SAM, cusp, reaging-gan, lifespan, hrfae}. However, these methods often face difficulties in capturing high-resolution details and ensuring identity preservation. These limitations can result in artifacts or inaccurate reconstructions, which restrict their applicability to broader and more diverse scenarios. Recently, diffusion models (DMs) have emerged as a powerful alternative, offering enhanced semantic understanding and visual fidelity using inversion techniques for image synthesis and editing~\cite{blended-diffusion-1, diffedit, p2p, null-text, li2023stylediffusion, negative-prompt, proximal, direct-inversion}. 

While modern DMs can produce detailed results, they often fail to capture complex, abstract conditions like face aging, where intrinsic and extrinsic factors must be disentangled for realistic synthesis. To address this, we introduce \textit{Face-Attribute-Aware Refinement} strategy that refines the prompts with identity-related and condition-driven attributes. This enables context-aware fine-tuning of DMs, leading to identity consistent, realistic face aging.

Furthermore, conventional methods~\cite{null-text, li2023stylediffusion} depend on iterative optimisation to align real images with the model’s latent space, which is computationally intensive and often results in subpar reconstructions that fail to retain essential facial details.
Recently, tuning-free inversion techniques~\cite{negative-prompt, proximal, direct-inversion} have been proposed to address these issues. However, they often sacrifice reconstruction accuracy or have difficulty in generalizing across diverse image distributions. Our work advances this field by introducing a robust tuning-free \textit{Angular Inversion} method. This approach ensures fast and high-fidelity mapping of images into the latent space, enabling efficient and precise facial edits suitable for real-time applications.
% Traditional methods~\cite{null-text, li2023stylediffusion} rely on iterative optimization to align real images with the model’s latent space, which is computationally expensive and often leads to suboptimal reconstructions that fail to preserve critical facial details. 

To leverage the benefits of \textit{Angular Inversion}, it is imperative to advance attention control mechanisms for more effective guided image editing. Existing methods~\cite{p2p, pnp, masactrl, FPE} are typically static, applying uniform strategies across the whole image, which creates artifacts and background hallucination due to their inability to isolate and prioritize age-relevant regions. Furthermore, static approaches lack the flexibility to adapt to the semantic importance of various facial attributes, which is crucial for realistic age transformations. To address this, we introduce \textit{Adaptive Attention Control}, a dynamic mechanism that modulates cross/self-attention during the editing process. 

Existing evaluation metrics for face aging primarily rely on comparing re-aged results with real target-age images. However, such reference-based evaluations are often unreliable due to the scarcity of paired ground-truth data, making consistent identity assessment difficult. To provide a reference-independent assessment of identity consistency across diverse age transitions, we propose a new evaluation protocol termed \textit{Cyclic Identity Similarity}, which evaluates the consistency of re-aged faces through cyclic transformations (e.g., 40→60→40), as illustrated in Figure~\ref{fig:intro-cylic}. 
% This metric offers a robust, reference-independent assessment of identity consistency across diverse age transitions.

To summarize, we propose a new face aging framework called FaceTT, where we introduce:
\begin{itemize}
    \item \textit{\textbf{Face-Attribute-Aware Prompt Refinement}} strategy that encodes intrinsic and extrinsic aging attributes to refine prompts, enabling context-aware conditioning for realistic age progression while preserving identity.
    % Leveraging FastVLM~\cite{vasu2025fastvlm}, we automatically extract facial attributes such as age, gender, skin tone, texture, and condition-related cues (e.g., hair loss, sleep deprivation, or weight gain) from a face image to construct descriptive prompts. 
    % These semantically rich prompts provide the diffusion model with precise, context-aware conditioning, enabling faithful reconstruction and realistic age progression while maintaining the individual’s identity.
    % \item A new cross-age expert framework called \textcolor{red}{FaceTT}, which introduces a novel \textcolor{red}{A}ngular In\textcolor{red}{v}ersion and \textcolor{red}{A}daptive A\textcolor{red}{t}tention Control for F\textcolor{red}{a}cial \textcolor{red}{R}e-Aging task.
    \item \textbf{\textit{Angular Inversion}}, a new tuning-free inversion technique that efficiently maps input images to the latent space of a pre-trained DM without iterative optimization and reduces computational overhead without sacrificing quality.
    % It separates source and target branches to enhance content preservation (source) and edit fidelity (target) in facial re-aging. 
    % It leverages angular differences between latent vectors and scales updates using cosine similarity for smooth, artifact-free edits. 
    % This approach reduces structural discrepancies without iterative optimization and reduces computational overhead enabling real-time editing without sacrificing quality.
    \item \textbf{\textit{Adaptive Attention Control} (\textit{AAC})} that determines when to apply cross-attention for semantic changes and when to use self-attention to maintain structural integrity of faces and allows precise control over aging effects while preserving identity and contextual features.
    % \item \textbf{\textit{Adaptive Attention Control} (\textit{AAC})} that leverages a KL-divergence-based threshold and determines when to apply cross-attention for semantic changes and when to use self-attention to maintain structural integrity of faces. This adaptive approach allows precise control over aging effects while preserving identity and contextual features.
    % \item We propose a new evaluation protocol, Cyclic Identity Similarity, which assesses aging and de-aging performance by computing identity consistency across cyclic age groups (e.g., 40→60→40), as shown in Figure~\ref{fig:intro-cylic}. This metric .
    \item \textbf{\textit{Cyclic Identity Similarity}}, a new evaluation protocol that measures identity preservation and visual realism by assessing consistency across cyclic age transformations.
    % \item Through extensive experiments, we show that our method outperforms state-of-the-art (SOTA) face aging methods. To effectively measures identity preservation and visual realism with respect to the reference image, we introduce a new evaluation protocol called \textit{\textbf{Cyclic Identity Similarity}} which assesses aging performance by computing identity consistency across cyclic age groups.
\end{itemize}
% Through extensive experiments, we show that our method outperforms state-of-the-art (SOTA) face aging methods.

\section{Related Works} \label{sec:related_work}

\noindent\textbf{Inversion and Editing in DMs:} 
Inversion and attention control are central for faithful, structure-preserving edits in DMs. DDIM inversion~\cite{p2p} offers deterministic mapping but often introduces reconstruction errors that lead to identity drift. Null-text inversion~\cite{null-text} enhances stability but increases computational overhead, whereas Negative-prompt inversion~\cite{negative-prompt} reduces computational overhead at the cost of image fidelity. 
ReNoise~\cite{renoise} enhances inversion robustness by mitigating error propagation. Direct inversion~\cite{direct-inversion} separates source and target branches to balance content preservation and edit fidelity. Complementing these, attention-based editing frameworks such as P2P~\cite{p2p}, PnP~\cite{pnp}, and MasaCtrl~\cite{masactrl} adjust attention maps to maintain spatial and semantic coherence. FPE~\cite{FPE} shows stable transformations, while Inversion-free editing~\cite{inversion-free} integrates cross- and self-attention control to handle both rigid and non-rigid changes effectively.

\noindent\textbf{Face Aging:}  
Prior face aging approaches mainly rely on conditional GANs, where age is provided as a control signal~\cite{reaging1, wasserstein/reaging2, continuous/reaging3, reaging4, reaging5}. Methods like HRFAE~\cite{hrfae} manipulate latent spaces, CUSP~\cite{cusp} disentangles style and content, and MyTimeMachine~\cite{qi2025mytimemachine} focuses on personalized temporal aging. However, inverting real faces into GAN latent spaces often causes identity distortion~\cite{tov2021designing}.
While biometric constraints~\cite{Idendity_biometric} can reduce drift, they require strong identity labels and may suppress noticeable age changes. 
Recently, FADING~\cite{FADING} introduced a diffusion-based model for age editing, but it relies on costly optimization-based inversion~\cite{null-text}. Multiverse Aging~\cite{gong2025agingmultiverse} explores condition-aware, multi-trajectory aging but trades off fine-grained reconstruction control. TimeBooth~\cite{timebooth} improves personalization but still struggles with subtle detail, background consistency, and fine-grained identity preservation.

Despite advancements, prior works struggle to ensure high-fidelity aging, background and identity preservation under varying internal and external aging conditions. 
Our work addresses this gap by introducing a tuning-free \textit{Angular Inversion}, an  \textit{Adaptive Attention Control} mechanism and a \textit{Face-Attribute-Aware Prompt Refinement} strategy for context-aware, identity-preserved aging.

% \begin{figure*}[t]
%     \centering
%     \includegraphics[width = \linewidth, height = 5.75cm]{images/Network_updated.png}
%     \vspace{-2em}
%     \caption{\textbf{Overview of the facial re-aging pipeline.} The left diagram illustrates the workflow, beginning with \textit{Angular Inversion}, which maps the input image into the latent space of the diffusion model for accurate, non-iterative reconstruction. This is followed by \textit{Adaptive Attention Control}, which uses a KL-divergence-based threshold to switch between cross-attention for age-specific edits (e.g., wrinkles) and self-attention for preserving structural consistency. Source and target prompts guide the edits, enabling precise, high-quality facial re-aging while maintaining essential features.}\label{fig:intro}
%     \vspace{-1em}
% \end{figure*}

\section{Face Aging Methodology -- FaceTT}
\begin{figure}[t]
    \centering
    \includegraphics[width = \linewidth]{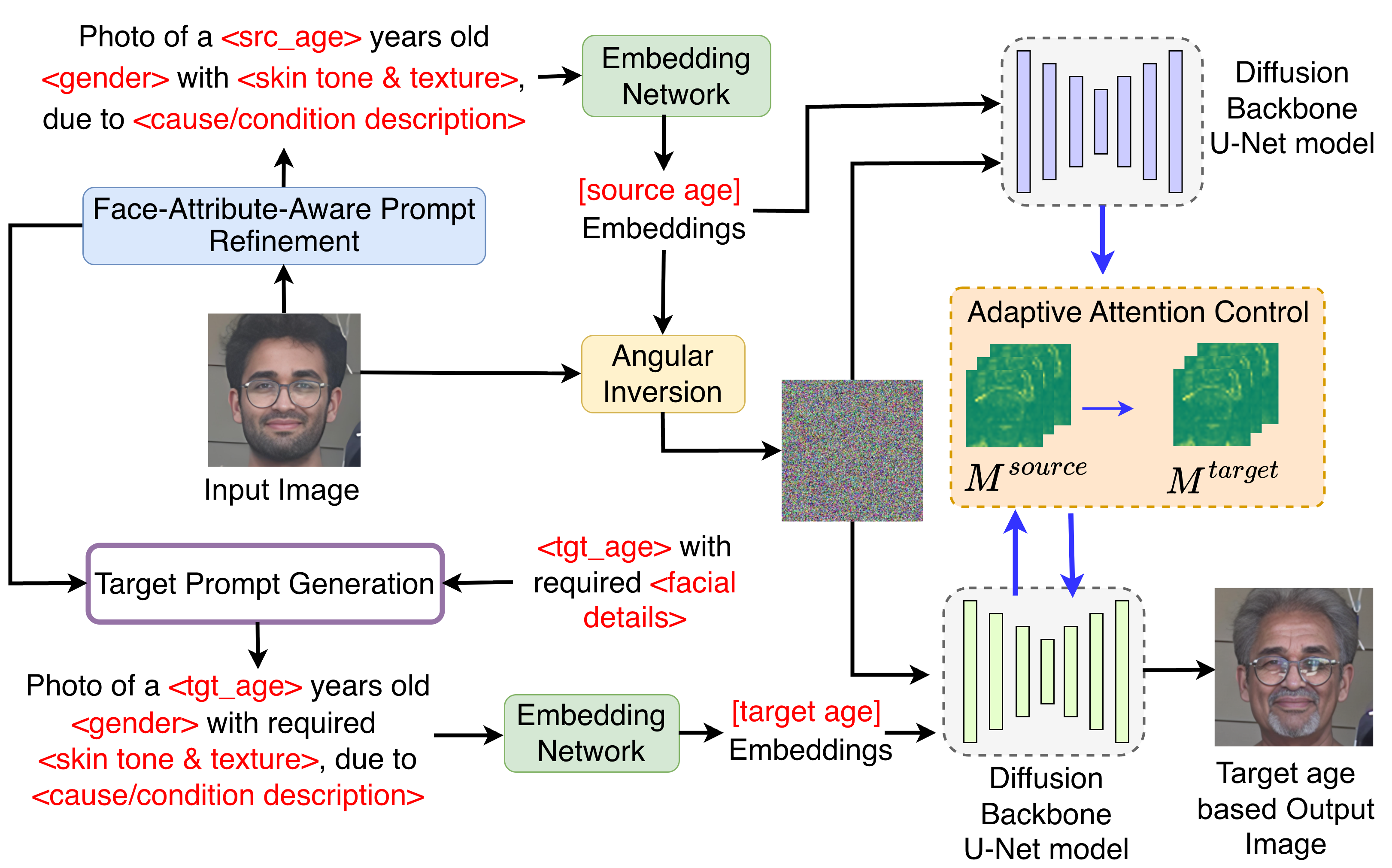}
    \vspace{-2em}
    \caption{Overview of the proposed FaceTT framework. Given an input face, the \textit{Face-Attribute-Aware Prompt Refinement} strategy produces a refined, attribute-rich facial prompts. These prompts are encoded using the embedding network, where the source age embeddings guide the \textit{Angular Inversion} process to obtain high-fidelity latent reconstructions. Both source and target age embeddings are then injected into the diffusion model through \textit{Adaptive Attention Control}, enabling precise and identity-consistent age editing. Finally, the age-transformed face is generated.
    % Given an input face, the \textit{Face-Attribute-Aware Prompt Refinement} module generates face detailed prompt. These refined prompts are encoded using the embedding network. The source age based embeddings are then guide the \textit{Angular Inversion} to obtain high-fidelity latent reconstructions. While source and target age based embeddings are injected into the diffusion model via \textit{Adaptive Attention Control}. Finally, the age-transformed face is generated.
    } \label{fig:main}
    \vspace{-1.5em}
\end{figure}

% \textbf{Problem statement:} 
% Let $I_{src}$ denote source image with the source prompt $C_{src}$ set to \texttt{Photo of a <src\_age> years old person}. Our objective is to synthesize a new desired image  $I_{tgt}$ with the target prompt $C_{tgt}$ set to \texttt{Photo of a <tgt\_age> years old person} while preserving the identity of the person in $I_{src}$. 
% Given an image with \texttt{<src\_age>} age, our objective is to synthesize a new image with any target age \texttt{<tgt\_age>}. Figure~\ref{fig:main} presents an overview of the proposed face aging framework, FaceTT. We use a pre-trained text-to-image diffusion model, and then initially fine-tune it using two types of prompts: an age-specific prompt \texttt{Photo of a <curr\_age> years old <person>} and an age-agnostic prompt \texttt{Photo of a <person>}. When finetuning for face aging task, describing skin tone and facial texture cues is crucial because the diffusion model learns subtle correlations that reflect lifestyle and environmental aging factors. Thus expanding the prompt accordingly is necessary, which is handled by \textit{Face-Attribute-Aware Prompt Refinement} strategy. Building on this textual conditioning, the inversion stage leverages the proposed \textit{Angular Inversion} technique for high-quality reconstructions. Age editing is then accomplished through the proposed \textit{Adaptive Attention Control} (\textit{AAC}) technique, which ensures that only age-relevant features are modified. 
Given an image with \texttt{<src\_age>}, our goal is to synthesize a new image at any target age \texttt{<tgt\_age>}. Figure~\ref{fig:main} illustrates the proposed FaceTT framework. We employ a pre-trained text-to-image diffusion model, initially fine-tuned using both age-specific prompts (\textit{Photo of a \texttt{<src\_age>} years old \texttt{<person>}}) and age-agnostic prompts (\textit{Photo of a \texttt{<person>}}). To capture fine-grained cues of lifestyle and environmental aging, we incorporate the \textit{Face-Attribute-Aware Prompt Refinement} strategy, which enriches prompts with intrinsic and extrinsic facial attributes. Building on this conditioning, the proposed \textit{Angular Inversion} enables high-fidelity latent reconstruction, while \textit{Adaptive Attention Control} (AAC) selectively modulates age-relevant features, ensuring precise and identity-consistent face aging.
% The \texttt{<person>} token in the prompt is replaced with either of \texttt{<man}, \texttt{woman}, \texttt{boy}, \texttt{girl>} based on gender and age. This proposed method allows precise control over the effects of aging while maintaining identity and context. 
% Next, we provide a brief background of DDIM, followed by details of our proposed techniques. 

\subsection{Face-Attribute-Aware Prompt Refinement}
While modern DMs can generate highly detailed images, they often struggle to interpret complex conditions. 
% Facial aging is especially challenging where intrinsic and extrinsic aging factors must be disentangled to produce realistic age changes without compromising identity.
% While modern DMs are capable of generating highly detailed outputs, they often struggle to interpret complex or abstract conditions. Facial aging, in particular, is an ill-posed problem where intrinsic and extrinsic factors represent two fundamentally different sources of visual change that must be disentangled to achieve realistic and identity-preserving results.
Fine-tuning the model with simple prompts like \texttt{Photo of a <src\_age> years old <person>} often fails to produce satisfactory results, as it does not explicitly describe the attributes that influence facial transformation across time. Intrinsic factors refer to internal, biological, and identity-related attributes that evolve naturally with age but are inherent to the individual — they define who the person is and how they age. In contrast, extrinsic factors capture external, lifestyle, or environmental influences that affect facial appearance but are not biologically bound — they originate from the person’s surroundings or habits.

To capture these nuances, we construct refined prompts in the form:
\textit{Photo of a \texttt{<src\_age>} years old \texttt{<gender>} with \texttt{<skin tone \& texture>}, due to \texttt{<cause/condition description>}}. 
Here, \texttt{skin tone \& texture} represents intrinsic factors, while \texttt{cause/condition description} reflects extrinsic influences.
% Here, \textit{skin tone \& texture} corresponds to intrinsic factors, while \textit{cause / condition description} represents extrinsic influences. 
% For example, \texttt{Photo of a 50 years old man with fair skin tone and a receding hairline, thinning hair on the crown, and visible scalp due to hair loss.} 
We extract \texttt{age}, \texttt{gender}, \texttt{skin tone} and \texttt{texture}, and \texttt{cause/condition description} from the input face using a vision language model FastVLM~\cite{vasu2025fastvlm}.
% , and then we fine-tune the end-to-end framework using both age-specific and age-agnostic.
% To extract \textit{age}, \textit{gender}, \textit{skin tone} and \textit{texture}, and \textit{cause/condition description} from a face image, we have utilized the  FastVLM~\cite{vasu2025fastvlm} model. The model is then fine-tuned using both age-specific prompts and age-agnostic prompts.

This refinement is crucial as high-level conditions such as “hair loss” or “weight gain” are not directly grounded in low-level visual features without sufficient contextual detail. Expanding such vague inputs into semantically rich, visually grounded descriptions enables the model to better align the generated appearance with the intended condition, thereby enhancing both realism and identity preservation.

\begin{figure}[t]
    \centering
    \includegraphics[width = \linewidth, height = 4.8cm]{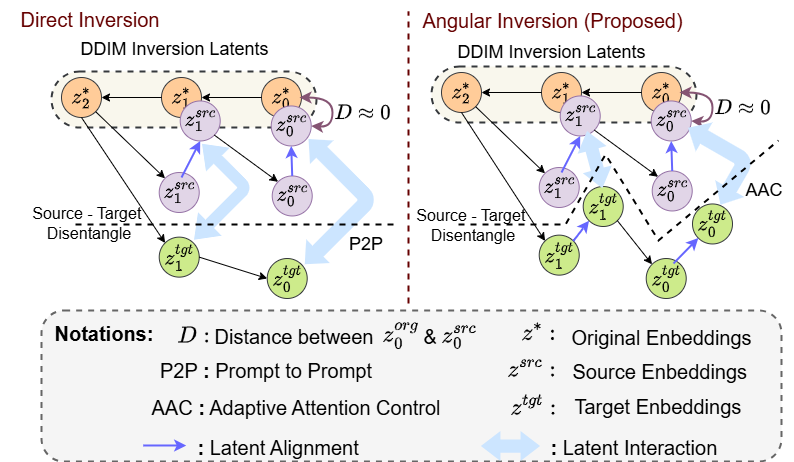}
    \vspace{-2em}
    \caption{Illustration of the Angular Inversion technique with prior inversion techniques.}\label{fig:angular}
    \vspace{-1em}
\end{figure}
\begin{algorithm}[t]
\caption{\small Angular Inversion}
\label{algorithm1}
\footnotesize
\begin{algorithmic}[1]
\Require A face-attribute-aware source prompt embedding $C^{src}$, a face-attribute-aware target prompt embedding $C^{tgt}$, a real image or latent embedding $z_0^{src}$, DI: DDIM Inversion function, DF: DDIM Forward function.
\Ensure An edited image or latent embedding $z_0^{tgt}$
\State \textbf{Part I : Inverse $z_0^{src}$}
\State $z_0^* = z_0^{src}$
\For{$t = 1, \dots, T - 1$}
    \State $z_t^* \gets \text{DI}(z_{t-1}^*, t - 1, C^{src})$
\EndFor
\State \textbf{Part II: Perform editing on $z_T^*$}
\State $z_T^{tgt} = z_T^*; z_T^{src} = z_T^*$
\For{$t = T, T - 1, \dots, 1$}
    \State $o_{t-1}^{src} \gets (z_{t-1}^* - \text{DF}_{src}(z_t^{src}, t, C^{src}))$
    \State $o_{t-1}^{tgt} \gets (z_{t-1}^* - \text{DF}_{tgt}(z_t^{tgt}, t, C^{tgt}))$ 
    % \Statex \hspace{1.5em} 
    \State $\theta_{t-1}^{src} = \arccos\left(\frac{(z_{t-1}^* - z_T^*) \cdot (z_{t-1}^{src} - z_T^*)}{\|z_{t-1}^* - z_T^*\| \|z_{t-1}^{src} - z_T^*\|}\right)$ \hfill
    \State $\theta_{t-1}^{tgt} = \arccos\left(\frac{(z_{t-1}^* - z_T^*) \cdot (z_{t-1}^{tgt} - z_T^*)}{\|z_{t-1}^* - z_T^*\| \|z_{t-1}^{tgt} - z_T^*\|}\right)$ \hfill
    % \State $\textcolor{red}{o_{t-1}^{tgt} \gets o_{t-1}^{tgt} \cdot \exp(- \alpha \cdot \theta_{t-1}^{tgt})}$ \hfill
    \State $z_{t-1}^{src} = \text{DF}_{src}(z_t^{src}, t, C^{src}) + o_{t-1}^{src}$ \hfill
    \State $o_{t-1}^{src} \gets o_{t-1}^{src} \cdot \exp(- \xi \cdot \theta_{t-1}^{src})$ \hfill
    \State $o_{t-1}^{tgt} \gets o_{t-1}^{tgt} \cdot \exp(- \xi \cdot \theta_{t-1}^{tgt})$ \hfill
    \State $\beta_{t-1} \gets \text{cosine\_similarity}(z_{t-1}^*, z_{t-1}^{tgt})$ \hfill
    \State $z_{t-1}^{tgt} \gets \text{DF}_{tgt}(z_t^{tgt}, t, C^{tgt}) +$
    \Statex \hspace{10em} $\beta_{t-1} \cdot o_{t-1}^{tgt} + (1 - \beta_{t-1}) \cdot o_{t-1}^{src}$
\EndFor
\State \Return $z_0^{tgt}$
\end{algorithmic}
\end{algorithm}
% \footnotetext{Will release complete code upon acceptance.}
% \vspace{-5pt}

\subsection{Angular Inversion Technique}
\noindent \textbf{Background \& Preliminaries:}
DMs iteratively transform noise \(z_T\) into an image \(z_0\) through Gaussian perturbations \(\epsilon \sim \mathcal{N}(0,1)\), with \(z_t = \sqrt{\alpha_t}z_0 + \sqrt{1-\alpha_t}\epsilon\). A denoiser \(\epsilon_\theta\) is trained to predict \(\epsilon(z_t, t)\). DDIM sampling~\cite{DDIM} generates \(z_{t-1}\) via the learned noise estimate.  
DDIM inversion reverses this process as 
\(z_t^* = \frac{\sqrt{\alpha_t}}{\sqrt{\alpha_{t-1}}}z_{t-1}^* + \sqrt{\alpha_t}\left(\sqrt{\frac{1}{\alpha_t}-1} - \sqrt{\frac{1}{\alpha_{t-1}}-1}\right)\epsilon_\theta(z_{t-1}^*, t-1)\),
though it often yields imperfect reconstruction. CFG~\cite{cfg} blends conditional and unconditional denoising, 
\(\epsilon_\theta(z_t,t,C,\emptyset) = w\epsilon_\theta(z_t,t,C) + (1-w)\epsilon_\theta(z_t,t,\emptyset)\), 
enhancing prompt ($C$) adherence but introducing latent inconsistencies between original and edited image's latent trajectories \(z_t^{src}\) and \(z_t^{tgt}\).

% \vspace{0.5em}
% \noindent \textbf{Existing Inversion Techniques:} 
To mitigate inconsistencies from DDIM inversion and CFG, optimization-based methods 
\cite{null-text, negative-prompt, proximal, direct-inversion, edit-friendly-ddpm} have been proposed. Null-Text Inversion~\cite{null-text} refines a null-text embedding to minimize \(z_t^{src/tgt} - z_t^*\), but is slow and prone to latent misalignment. Direct Inversion~\cite{direct-inversion} improves content preservation and edit fidelity by decoupling source and target branches, yet struggles with complex, non-linear edits and may underperform when substantial semantic changes (wide age transformations) are required.

To address the limitations of existing inversion techniques, we propose \textit{Angular Inversion}, which disentangles the source and target branches. This separation allows each branch to be optimized independently (see Figure~\ref{fig:angular}). In the source branch, we directly add \((z_t^* - z_t^{src})\) to \(z_t^{src}\). In the target branch, rather than leaving it unaltered, we scale \((z_t^* - z_t^{src})\) by \(\exp(-\xi \cdot \angle z_t^* z_T^* z_t^{src})\) and \((z_t^* - z_t^{tgt})\) by \(\exp(-\xi \cdot \angle z_t^* z_T^* z_t^{tgt})\), where \(\xi\) is a hyper-parameter.

During iteration, large angles indicate misalignment. To mitigate this, we downweight updates proportionally, enabling gradual adjustments and preventing disruptive shifts in latent representations. This stabilizes the target branch and avoids errors during stepwise generation.

Next, a linear combination of the scaled \((z_t^* - z_t^{src})\) and \((z_t^* - z_t^{tgt})\), weighted by the cosine similarity between \(z_t^*\) and \(z_t^{tgt}\), is added back to \(z_t^{tgt}\). High similarity shifts focus toward \((z_t^* - z_t^{tgt})\) to improve edit fidelity, while low similarity increases the influence of \((z_t^* - z_t^{src})\).

This method enables efficient, high-quality edits by (1) removing costly optimization, (2) reducing discrepancies between source and target latent spaces, (3) improving target edit fidelity through better alignment, and (4) preserving the diffusion model’s input distribution. A detailed description is provided in Algorithm~\ref{algorithm1}.

\subsection{Adaptive Attention Control}
We investigate the distinct roles of cross- and self-attention in face aging using Stable Diffusion to understand their impact on identity-consistent age transformation (see \texttt{supplementary-S1}). Our analysis shows that cross-attention primarily captures semantic age-related cues—such as wrinkles, skin tone, and hair color—driven by the conditioning text. In contrast self-attention preserves facial geometry, expression, and identity coherence. This probing study enables fine-grained manipulation of facial features by selectively modifying cross-attention maps, allowing controlled semantic aging while retaining the original face layout. However, applying cross-attention control throughout the full sampling process tends to over-constrain spatial alignment, limiting the model’s flexibility to generate realistic age variations. Hence, following~\cite{p2p}, we restrict cross-attention control to the early denoising steps (before $\tau_1$), ensuring a balanced trade-off between semantic transformation and structural preservation.

% A key limitation of cross-attention control in face aging lies in its inability to model non-rigid facial deformations that naturally occur with aging. As observed in~\cite{masactrl, FPE}, such flexible spatial adaptations are instead governed by self-attention, where query maps preserve the overall facial layout while accommodating semantic variations required by the target age prompt. Our probing analysis (see \texttt{supplementary-S1}) confirms this behavior. Following~\cite{FPE}, we replace the self-attention maps of layers 4–14 in the target image with those from the source, but only after $\tau_2$ steps. This ensures that non-rigid yet identity-consistent age transformations are achieved without introducing distortions or inconsistencies in facial composition—particularly around accessories such as glasses, hats, or earrings, and in complex background regions.
\begin{figure}[t]
    \centering
    \includegraphics[width = \linewidth, height = 4.8cm]{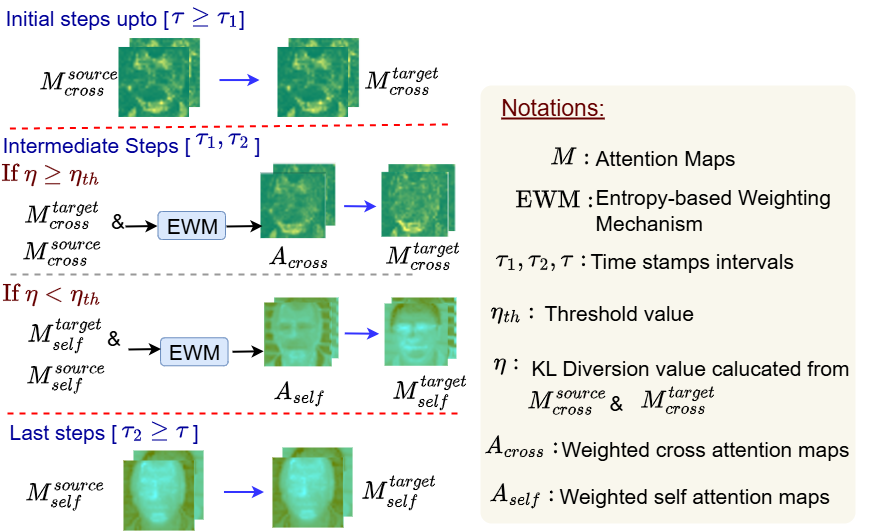}
    \vspace{-1.5em}
    \caption{Illustration of Adaptive Attention Control Mechanism.}\label{fig:AAC}
    \vspace{-1em}
\end{figure}
\begin{algorithm}[t]
\caption{\small Adaptive Attention Control}
\label{algorithm2: AAC}
\footnotesize
\begin{algorithmic}[1]
\Require Face-attribute-aware source prompt embedding $C^{src}$, face-attribute-aware target prompt embedding $C^{tgt}$, real image or latent embedding $z_0^{src}$, H: entropy function, DI: DDIM Inversion function, DF: DDIM Forward function.
\Ensure An edited image or latent embedding $z_0^{tgt}$
\State \textbf{Part I : Inverse $z_0^{src}$}
\State $z_0^* = z_0^{src}$
\For{$t = 1, \dots, T - 1$}
    \State $z_t^* \gets \text{DI}(z_{t-1}^*, t - 1, C^{src})$
\EndFor
\State \textbf{Part II: Perform editing on $z_T^*$}
\State $z_T^{tgt} = z_T^*; z_T^{src} = z_T^*$
\For{$t = T, T - 1, \ldots, 1$}
    \State $z_{t-1}^{src}, M_{\text{self}}^{\text{src}}, M_{\text{cross}}^{\text{src}} \gets \text{DF}_{src}(z_t^{src}, t, C^{src}) $
    \State $M_{\text{self}}^{\text{tgt}}, M_{\text{cross}}^{\text{tgt}} \gets \text{DF}_{tgt}(z_t^{tgt}, t, C^{tgt}) $ % FIX 1: added opening $
    \If{$t > \tau_1$}
        \State $z_{t-1}^{tgt} \gets \text{DF}_{tgt}(z_t^{tgt}, t, C^{tgt})\{M_{\text{cross}}^{tgt} \leftarrow M_{\text{cross}}^{src}\}$
    \ElsIf{$\tau_2 \leq t \leq \tau_1$}
        \State $\eta \gets \text{D}_\text{KL}({M_{\text{cross}}^{src}} || {M_{\text{cross}}^{tgt})}$ % FIX 2: added opening $
        \If{$\eta > \eta_{th}$} % FIX 3: removed extra }
            \State $w_{t-1} \gets 1 - \text{H}(M_{\text{cross}}^{src})$
            \State $A_{\text{cross}} \gets w_{t-1} \cdot M_{\text{cross}}^{src} + (1 - w_{t-1}) \cdot {M_{\text{cross}}^{tgt}}$ % FIX 4: added closing }
            \State $z_{t-1}^{tgt} \gets \text{DF}_{tgt}(z_t^{tgt}, t, C^{tgt})\{M_{\text{cross}}^{tgt} \leftarrow A_{\text{cross}}\}$
        \Else
            \State $w_{t-1} \gets 1 - \text{H}(M_{\text{self}}^{src})$
            \State $A_{\text{self}} \gets w_{t-1} \cdot M_{\text{self}}^{src} + (1 - w_{t-1}) \cdot {M_{\text{self}}^{tgt}}$ % FIX 5: added closing }
            \State $z_{t-1}^{tgt} \gets \text{DF}_{tgt}(z_t^{tgt}, t, C^{tgt})\{M_{\text{self}}^{tgt} \leftarrow A_{\text{self}}\}$
        \EndIf
    \Else
        \State $z_{t-1}^{tgt} \gets \text{DF}_{tgt}(z_t^{tgt}, t, C^{tgt})\{M_{\text{self}}^{tgt} \leftarrow M_{\text{self}}^{src}\}$
    \EndIf
\EndFor
\State \Return $z_0^{tgt}$
\end{algorithmic}
\end{algorithm}
A key limitation of cross-attention control in face aging is its inability to capture non-rigid facial deformations that naturally occur with aging. As observed in~\cite{masactrl, FPE}, such flexible spatial adaptations are better handled by self-attention, where query maps preserve facial layout while accommodating age-related semantic variations. Our probing analysis (see \texttt{supplementary-S1}) confirms this behavior. Following~\cite{FPE}, we replace the self-attention maps of layers 4–14 in the target image with those from the source, but only after $\tau_2$ steps. This design enables non-rigid yet identity-consistent transformations without introducing distortions—particularly around accessories (e.g., glasses, hats, earrings) and in complex background regions.

% During the intermediate steps, i.e., between $\tau_1$ and $\tau_2$, enabling rigid and non-rigid semantic changes presents challenges. Authors in~\cite{inversion-free} demonstrated that naively combining cross-attention and mutual self-attention can yield sub-optimal results in a dual-branch framework. Consequently, the attention control parameter (i.e., $\eta$) is determined by the KL-divergence between $M_{cross}^{src}$ and $M_{cross}^{tgt}$ (cross attentions of source and target branch). If $\eta$ falls below a specific threshold ($\eta_{th}$), self-attention is favored to preserve subtle structural features, such as the shape of the eyes, nose, or jawline, thereby maintaining the original identity of a person while making slight adjustments to age-related attributes. On the other hand, if $\eta$ exceeds $\eta_{th}$, indicating significant differences between the cross-attention maps, the algorithm shifts to cross-attention to facilitate more substantial changes like adding deep wrinkles, or altering facial features to create realistic aging effects.

During the intermediate denoising phase ($\tau_1$–$\tau_2$), achieving a balance between rigid and non-rigid semantic changes is challenging. As noted in~\cite{inversion-free}, naive fusion of cross- and self-attention often leads to sub-optimal edits in dual-branch frameworks. To address this, we dynamically regulate the attention control parameter $\eta$ based on the KL-divergence between the source and target cross-attention maps ($M_{cross}^{src}$ and $M_{cross}^{tgt}$). When $\eta$ is below a defined threshold ($\eta_{th}$), self-attention dominates to preserve fine structural details—such as the eyes, nose, and jawline—ensuring identity stability. Conversely, when $\eta > \eta_{th}$, the model prioritizes cross-attention to introduce more pronounced semantic transformations, such as deep wrinkles or other age-specific facial modifications.

% Instead of directly replacing target attention with source attention, the intermediate step blends them to ensure a smooth transition and avoid abrupt changes. This blending uses a weighting factor $w_t \gets 1 - \text{H}(M_{\text{cross}}^{src/tgt})$, determining the influence of source attention maps (whether self-attention or cross-attention) compared to the target attention maps. Here, H denotes the entropy of the source attention map which gives insight into how dispersed or concentrated the attention is. Low entropy makes $w_t$ close to 1, indicating focused attention that suggests certain structural elements of the source image are crucial. We assign more weight to these features to maintain the integrity of the original image. Conversely, high entropy makes $w_t$ close to 0, indicating dispersed attention and allowing more flexibility for incorporating target features, thereby facilitating non-rigid changes like age-related transformations. A similar analysis applies to self-attention maps.
% This process is detailed in Algorithm \ref{algorithm2: AAC} and Figure~\ref{fig:AAC}.

Instead of directly replacing target attention maps with those of the source, the intermediate stage blends them to ensure a smooth transition and prevent abrupt changes. The blending weight is defined as $w_t \gets 1 - \text{H}(M_{\text{cross}}^{src/tgt})$, which determines the relative influence of the source (either self- or cross-attention) and target attention maps. Here, $\text{H}(\cdot)$ denotes the entropy of the source attention map, reflecting how focused or dispersed the attention distribution is. Low entropy (i.e., concentrated attention) yields $w_t \approx 1$, emphasizing key structural elements of the source image to preserve identity. Conversely, high entropy produces $w_t \approx 0$, allowing greater flexibility for integrating target-specific attributes and enabling non-rigid, age-related transformations. A similar weighting scheme is applied to self-attention maps. The full procedure is outlined in Algorithm~\ref{algorithm2: AAC} and visualized in Figure~\ref{fig:AAC}.

\section{Experimental Analysis}
\vspace{-0.5em}
\textbf{Implementation details:} 
We leverage the pre-trained Stable Diffusion model~\cite{StableDiffusion} and finetune it on the FFHQ-Aging dataset~\cite{lifespan} for 150 steps with a batch size of 2. We incorporate the central age of the true age group into the finetuning prompt as \texttt{<src\_age>}. We employ Adam optimizer with a learning rate of $5 \times 10^{-6}$. For inversion, we set $\xi$ to 1.2. During attention control, $\eta_{th}$ is fixed at 0.05, while $\tau_1$ and $\tau_2$ are set to 35 and 15, respectively. 
All experiments are conducted on a single A100 GPU. We compare our model against SOTA face aging methods, including HRFAE~\cite{hrfae}, CUSP~\cite{cusp} and FADING~\cite{FADING}, with results reproduced using their official repos for fair comparison.
% We also include comparisons with recent unpublished works, Aging Multiverse~\cite{gong2025agingmultiverse}\footnote{https://agingmultiverse.github.io/} and MyTimeMachine (MyTM)~\cite{qi2025mytimemachine}\footnote{https://mytimemachine.github.io/} using the results provided on their official webpages.

\noindent\textbf{Dataset details:} 
We use two benchmark datasets, CelebA-HQ~\cite{karras2019style} and FFHQ-Aging~\cite{lifespan}, along with an in-the-wild celebrity age progression test set. CelebA-HQ is used only for evaluation. Following prior works~\cite{hrfae, reaging5}, age labels are obtained using the DEX classifier~\cite{dex}. We sample 1,000 “young’’ images and age them to 60, with all images resized to $512\times512$. FFHQ-Aging consists of 70,000 images at $1024\times1024$ resolution, grouped into ten age ranges from 0–2 to 70+ years.
For the in-the-wild evaluation, we construct a celebrity age progression test set by curating five celebrities with publicly available, license-compliant images across three age ranges: \{20–40\}, \{40–60\} and \{60–80\} years. One representative image per range yields a total of 15 images, enabling both reference-based and cyclic evaluations of aging quality and identity consistency.

\noindent \textbf{Cyclic Identity Similarity Protocol:} 
Traditional evaluations rely on comparing re-aged images with real target-age references. However, such reference-based comparisons are often unreliable due to limited ground-truth availability and high inter-age variations in real datasets. To address this limitation, we propose \textit{Cyclic Identity Similarity}, a new evaluation protocol that eliminates dependence on external ground-truth images. It consists of two analysis: \textit{Cyclic Identity Similarity} and \textit{Reference-Based Identity Similarity}.
% self-consistency-based metric 

% To evaluate identity preservation in face aging, we introduce a \textit{Cyclic Identity Similarity} protocol consisting of two metrics: \textit{Cyclic Identity Similarity} and \textit{Reference-Based Identity Similarity}.
\begin{enumerate}[leftmargin=*]
\item \textbf{Cyclic Identity Similarity ($\text{ID}_{sim}^{cyc}$):}  
% $\text{ID}_{sim}^{cyc}$ measures whether an aged face, when reverted to its original age, retains the same identity representation—providing a robust, dataset-independent evaluation of identity stability across transformations. 
Given an input image $I_{20\text{--}40}$ from the celebrity testset, we first age it to a target age range, producing $I'_{40\text{--}60}$, and then revert the process to reconstruct $I''_{20\text{--}40}$. Identity preservation across this round-trip transformation is measured as $
\text{ID}_{sim}^{cyc}(I_{20\text{--}40}) = \text{sim}\big(f(I_{20\text{--}40}),\, f(I''_{20\text{--}40})\big),$
where $f(\cdot)$ is a pretrained ArcFace~\cite{deng2019arcface} model. This assesses identity consistency over cyclic aging paths ($\{20\text{--}40\}\rightarrow\{40\text{--}60\}\rightarrow\{20\text{--}40\}$).

\item \textbf{Reference-Based Identity Similarity ($\text{ID}_{sim}^{ref}$):}  
Following~\cite{qi2025mytimemachine}, in the case of celebrity testset, for each re-aged image $I'_{a}$ (where $a \in \{\{40\text{--}60\},\,\{60\text{--}80\}\}$), we compare it to the corresponding real image $I_{a}$ of the same individual in the target age group:
$\text{ID}_{sim}^{ref}(I_{a}) = \text{sim}\big(f(I_{a}),\, f(I'_{a})\big).$
This evaluates how closely the synthesized aged face matches the true target-age identity.
\end{enumerate}

\noindent The averaged $\text{ID}_{sim}^{cyc}$ and $\text{ID}_{sim}^{ref}$ across all subjects serve as our primary quantitative measures.

\noindent \textbf{Additional Evaluation Metrics:}
Following~\cite{FADING}, we evaluate the face aging methods on 1) Mean Absolute Error (MAE), which compares the predictions of an age estimator to the target age; (2) Gender, Smile and Facial Expression Preservation, which measures the retention percentage of these attributes after transformation; (3) Blurriness, which assesses the degree of a facial blur; (4) Kernel-Inception Distance (KID), which quantifies the distance between real and generated images within the corresponding age groups. We employ Face++\footnote{faceplusplus.com/} to measure aging accuracy, attribute preservation and blurriness.

% Following~\cite{FADING}; we access the models using several metrics: (1) Mean Absolute Error (MAE), which compares the predictions of an age estimator to the target age; (2) Gender, Smile and Facial Expression Preservation, which measures the retention percentage of these attributes after transformation; (3) Blurriness, which assesses the degree of a facial blur; (4) Kernel-Inception Distance (KID), which quantifies the distance between real and generated images within the corresponding age groups. We employ Face++\footnote{https://www.faceplusplus.com/} to measure aging accuracy, attribute preservation and blurriness.

\subsection{Comparison with SOTA methods}
\textbf{Face Aging methods:} Table~\ref{tab:table4} presents a comparison on CelebA-HQ dataset. Note that a 5.14-year discrepancy (i.e., 65.14; first row) is reported for our training subset (age label 60) between the DEX classifier used for inference and the Face++ classifier used for evaluation. The results of the proposed FaceTT model are closer to the target age of 60. Additionally, it obtains highest score in gender preservation, demonstrating its effectiveness in maintaining age-independent characteristics. 
\begin{table}[t]
\centering
\caption{Quantitative results on CelebA-HQ dataset on the young-to-60 task. Best score: in \colorbox{red!20}{red}, second-best score: in \colorbox{celadon!50}{green}.} \label{tab:table4}
\vspace{-1em}
\begin{adjustbox}{width=0.45\textwidth}
\begin{tabular}{lcccccc}
\toprule
\textbf{Method} & \textbf{Predicted age} & \textbf{Blur $\downarrow$} & \textbf{Gender $\uparrow$} & \textbf{Smiling $\uparrow$} & \textbf{Neutral $\uparrow$} & \textbf{Happy $\uparrow$} \\ \midrule
Real images     & 65.14 $\pm$ 4.86 & 1.73 & ---  & ---  & --- & --- \\
HRFAE           & 55.05 $\pm$ 9.18 & \colorbox{celadon!50}{3.42} & 94.80 & 74.60 & \colorbox{celadon!50}{61.15} & 71.09 \\
CUSP            & 57.57 $\pm$ 7.88 & 3.39 & 89.79 & 75.88 & 60.67 & 70.87 \\
FADING          & \colorbox{celadon!50}{69.88 $\pm$ 6.20} & \colorbox{red!20}{2.18} & \colorbox{celadon!50}{98.44} & \colorbox{celadon!50}{76.17} & 56.54 & \colorbox{red!20}{73.19} \\
% \rowcolor{lightmintbg} 
\textbf{FaceTT}   & \colorbox{red!20}{62.05 $\pm$ 6.81} & \colorbox{red!20}{2.18} & \colorbox{red!20}{99.79} & \colorbox{red!20}{78.31} & \colorbox{red!20}{62.67} &  \colorbox{celadon!50}{72.17} \\ \bottomrule 
\end{tabular}
\end{adjustbox}
\vspace{-0.5em}
\end{table}
\begin{table}[t]
\centering
\caption{Quantitative results on different age groups of FFHQ-Aging dataset. Best score: in \colorbox{red!20}{red}, second-best score: in \colorbox{celadon!50}{green}. } \label{tab:table5}
\vspace{-1em}
\small
\begin{adjustbox}{width=0.47\textwidth}
\begin{tabular}{lccccccccccc}
\toprule
\textbf{Method} & \textbf{0-2} & \textbf{3-6} & \textbf{7-9} & \textbf{10-14} & \textbf{15-19} & \textbf{20-29} & \textbf{30-39} & \textbf{40-49} & \textbf{50-69} & \textbf{70+} & \textbf{Mean} \\ \midrule
\multicolumn{12}{l}{Mean Absolute Error - MAE (lower is better)} \\ \midrule
HRFAE            & 35.43        & 32.49        & 36.43        & 25.95          & 21.35           & \colorbox{red!20}{7.43}         & \colorbox{red!20}{8.75}         & \colorbox{celadon!50}{13.72}         & 15.75         & 21.13         & 21.84         \\
CUSP            & 19.39        & 23.96        & 21.96        & 14.21          & 9.81           & 13.11         & 14.09         & 15.86         & 16.09         & 15.56         & 16.40         \\
FADING          & \colorbox{celadon!50}{11.63}        & \colorbox{celadon!50}{16.97}        & \colorbox{celadon!50}{18.76}        & \colorbox{celadon!50}{13.70}          & \colorbox{red!20}{7.03}           & 11.18         & 13.98         & 15.45         & \colorbox{celadon!50}{14.15}         & \colorbox{celadon!50}{11.91}         & \colorbox{celadon!50}{13.47}         \\
% Ours            & 11.12        & 18.89        & 14.01        & 16.85          & 8.10           & 8.92          & 14.10         & 15.42         & 12.98         & 11.04         & 12.04         \\ \hline
% \rowcolor{lightmintbg} 
\textbf{FaceTT}   & \colorbox{red!20}{11.05}        & \colorbox{red!20}{16.24}        & \colorbox{red!20}{12.45}        & \colorbox{red!20}{12.31}          & \colorbox{celadon!50}{7.10}           & \colorbox{celadon!50}{8.92}          & \colorbox{celadon!50}{11.23}         & \colorbox{red!20}{12.31}         & \colorbox{red!20}{11.12}        & \colorbox{red!20}{11.35}         & \colorbox{red!20}{11.40}         \\ \midrule
\multicolumn{12}{l}{Gender Accuracy (higher is better)} \\ \midrule
HRFAE            &  0.48       &     0.52    &   0.41      &     0.42      &      0.28      &   0.40       &    0.50      &  0.42        &   \colorbox{celadon!50}{0.52}       &   \colorbox{celadon!50}{0.55}       &   0.45       \\
CUSP            &  0.47       &   0.56      &  \colorbox{celadon!50}{0.60}      &   \colorbox{celadon!50}{0.58}        &     0.50       &      0.51    &    \colorbox{celadon!50}{0.43}      &   0.45       &    0.49      &     0.53     &     0.51     \\
FADING            &  \colorbox{celadon!50}{0.51}       &    \colorbox{celadon!50}{0.61}     &    0.57     &   0.53        &    \colorbox{celadon!50}{0.53}        &  \colorbox{celadon!50}{0.52}        &   0.40       &   \colorbox{celadon!50}{0.47}      &    \colorbox{celadon!50}{0.52}      &     \colorbox{celadon!50}{0.55}     &     \colorbox{celadon!50}{0.57}     \\
% \rowcolor{lightmintbg} 
\textbf{FaceTT}            &    \colorbox{red!20}{0.55}     &    \colorbox{red!20}{0.63}     &   \colorbox{red!20}{0.61}      &   \colorbox{red!20}{0.65}        &    \colorbox{red!20}{0.71}        &      \colorbox{red!20}{0.59}   &    \colorbox{red!20}{0.45}      &      \colorbox{red!20}{0.65}    &   \colorbox{red!20}{0.71}       &      \colorbox{red!20}{0.63}     &  \colorbox{red!20}{0.62}      \\ \midrule
\multicolumn{12}{l}{Kernel Inspection Distance - KID (lower is better)} \\ \midrule
HRFAE            & \colorbox{red!20}{0.37}     &   \colorbox{red!20}{0.36}      &    \colorbox{red!20}{0.34}     &  \colorbox{celadon!50}{0.49}        &    \colorbox{celadon!50}{0.37}        &        \colorbox{celadon!50}{0.31} &      \colorbox{celadon!50}{0.19}    &    0.35      &    \colorbox{celadon!50}{0.40}      &   \colorbox{celadon!50}{0.60}       &   \colorbox{red!20}{0.34}       \\
CUSP            &  13.10       &     7.20    &    3.70     &    1.60       &    0.70        &  0.41        &     0.34     &     0.31     &  0.53        &   2.70      &     3.06     \\
FADING            &   \colorbox{celadon!50}{10.32}      &   4.50      &   2.20      &   1.09        &  0.40          &    0.35      &     0.40     &    \colorbox{red!20}{0.28}      &  0.60        &   1.60       &    2.03      \\
% \rowcolor{lightmintbg} 
\textbf{FaceTT}            & 12.18        &   \colorbox{celadon!50}{1.03}      &   \colorbox{celadon!50}{0.76}      &  \colorbox{red!20}{0.39}       &     \colorbox{red!20}{0.36}       &     \colorbox{red!20}{0.30}   & \colorbox{red!20}{0.18}     &     \colorbox{celadon!50}{0.30}     &    \colorbox{red!20}{0.17}      & \colorbox{red!20}{0.20}         &     \colorbox{celadon!50}{1.58}     \\ \bottomrule
\hline
\end{tabular}
\end{adjustbox}
\vspace{-1em}
\end{table}
% \begin{figure}[t]
%     \centering
%     \includegraphics[width=0.47\textwidth]{wacv-2026-author-kit-template/images/Fig_6 (1).png}
%     \vspace{-1em}
%     \caption{Visual comparison of facial re-aging SOTA and our proposed methods. 
%     Our method achieves realistic aging effects with fewer artifacts, higher visual fidelity, and more consistent handling of complex facial features with significantly less inference time.}
%     \label{fig:visual-1}
%     \vspace{-1em}
% \end{figure}
% \end{figure}
\begin{figure}[t]
    \centering
    \includegraphics[width=0.475\textwidth, height = 6cm]{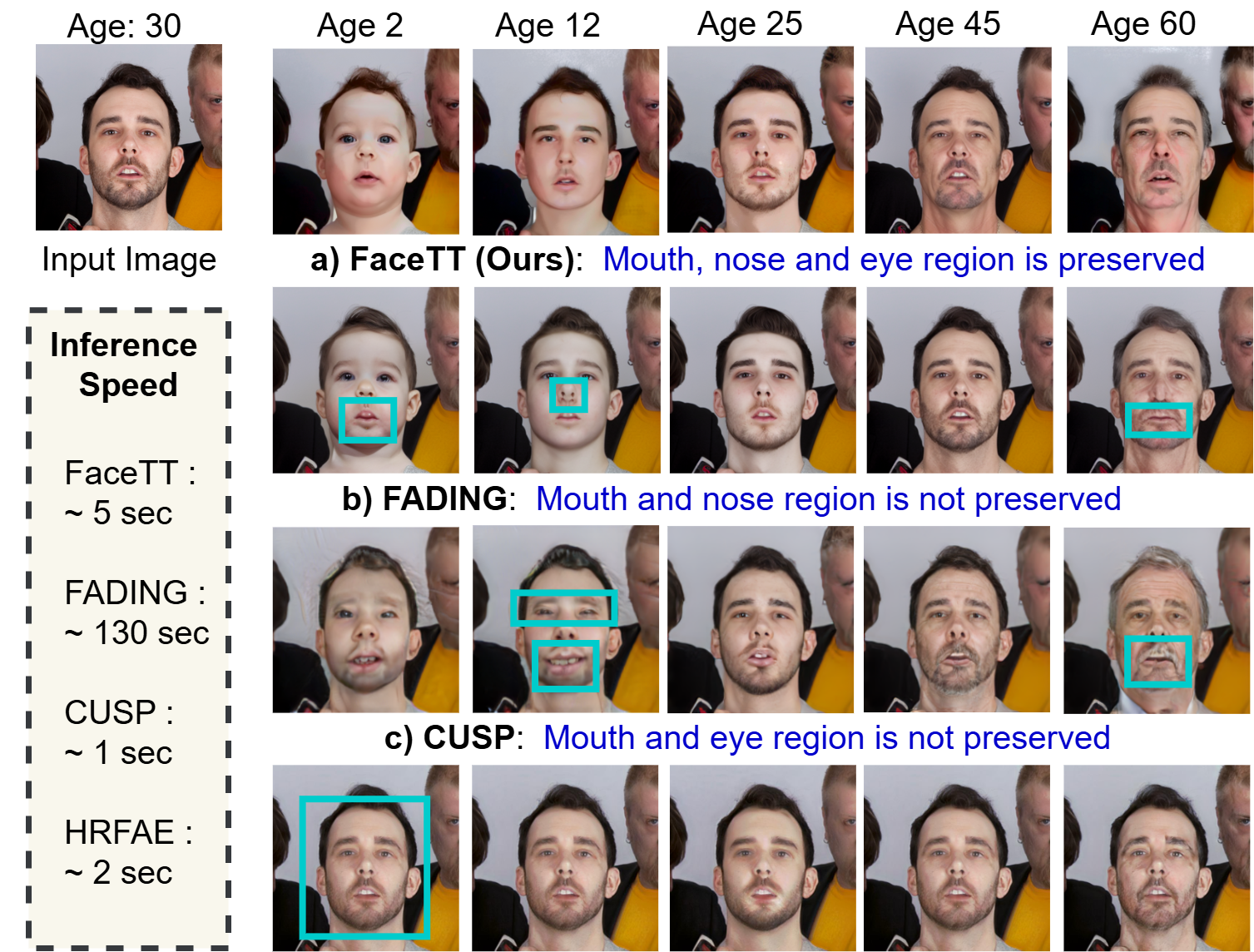}
    \vspace{-1.7em}
    \caption{Visual comparison with SOTA face aging methods.}
    \label{fig:visual-1}
    \vspace{-1em}
\end{figure}

\begin{figure}[t!]
    \centering 
    \includegraphics[width=0.475\textwidth]{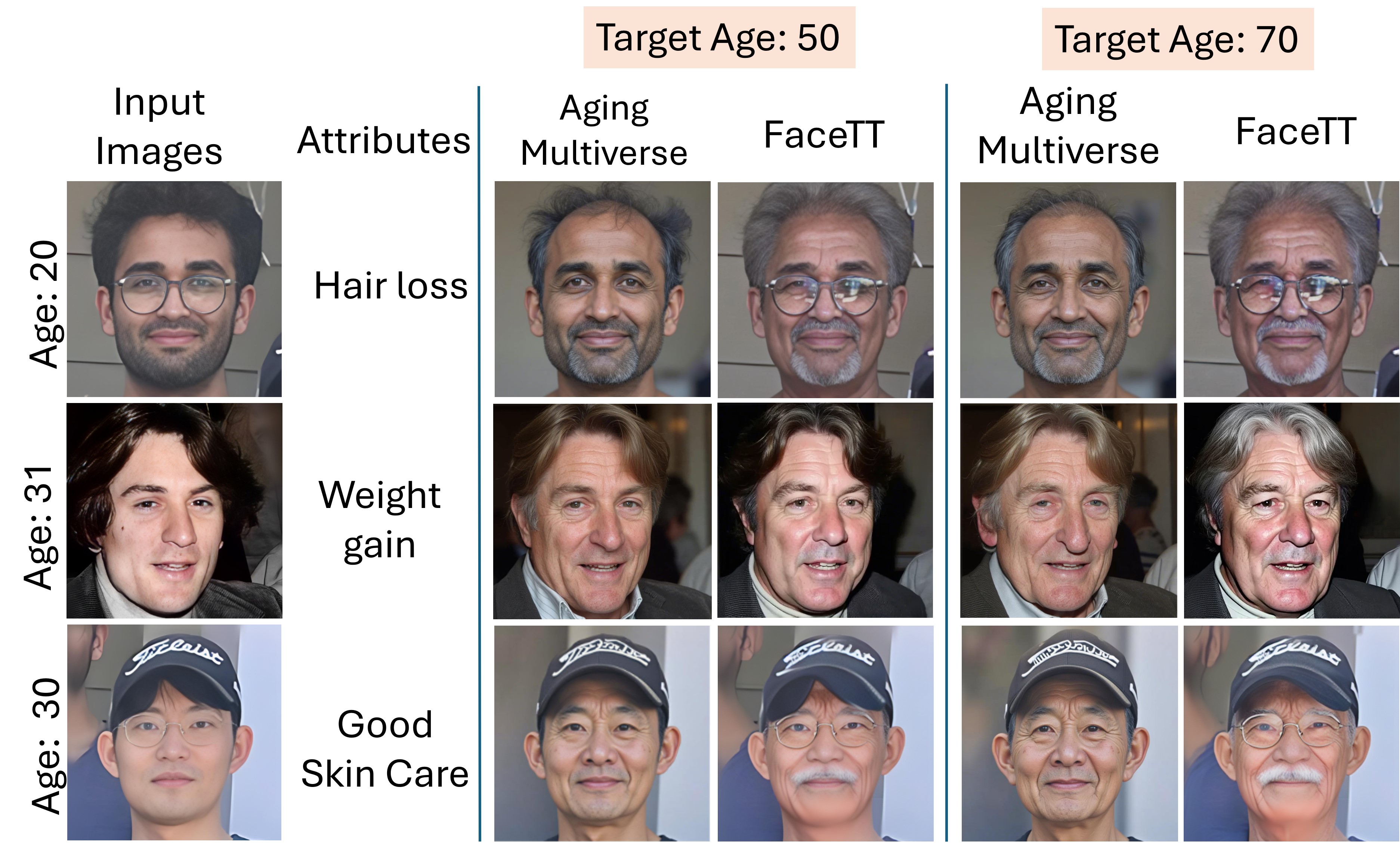}
    \vspace{-1.75em}
    \caption{Visual comparison with Aging Multiverse \cite{gong2025agingmultiverse}. Our method generates better attribute-aware aging while preserving background details (see rows 1 \& 2), skin tone (see rows 2 \& 3) and accessories such as eyeglasses (see rows 1 \& 3).}
    \label{fig:comparison_aging_multiverse}
    \vspace{-1em}
\end{figure}

\begin{figure}[t!]
    \centering 
    \includegraphics[width=0.46\textwidth]{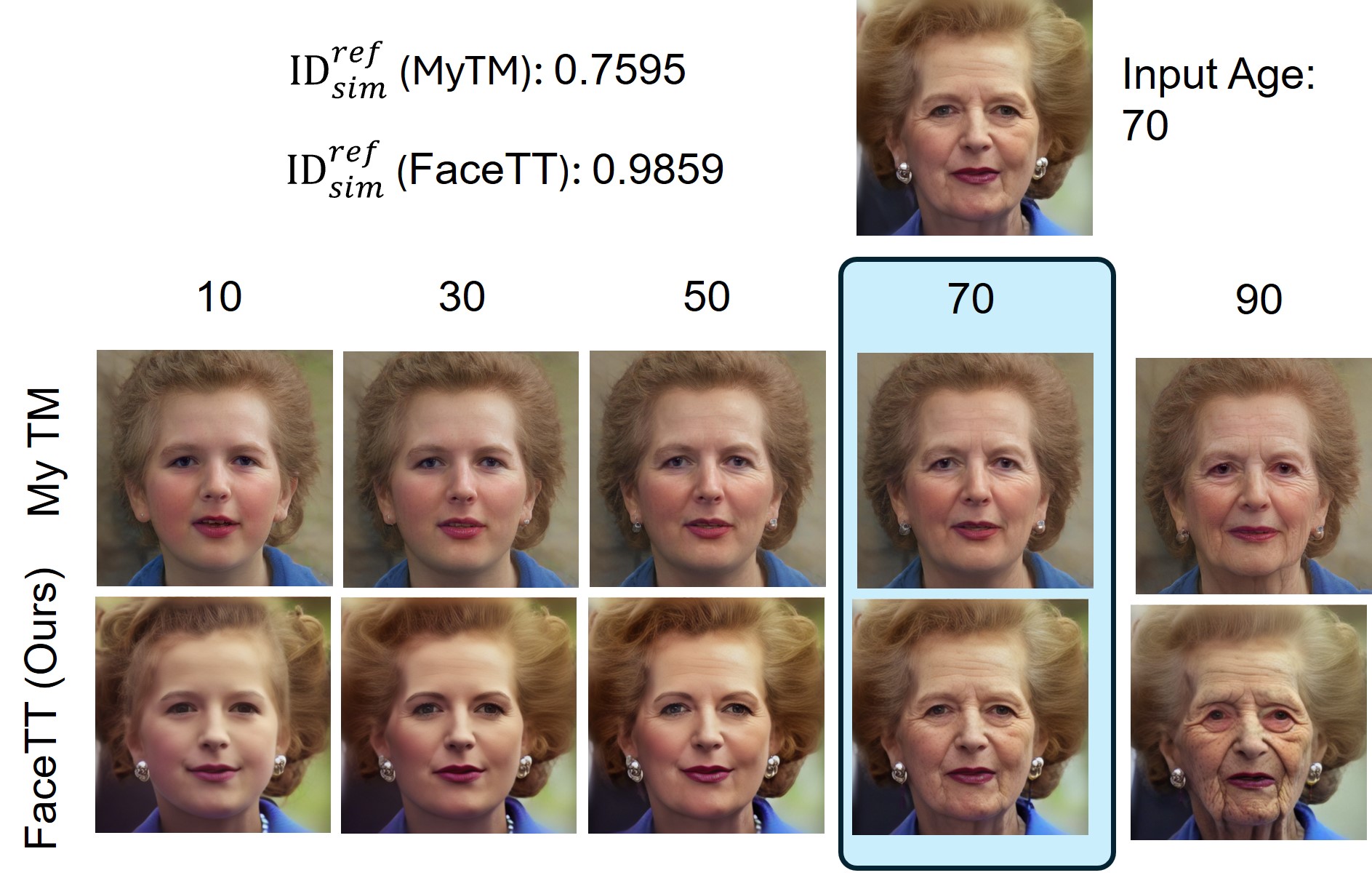}
    \vspace{-1em}
    \caption{Visual comparison with MyTM \cite{qi2025mytimemachine}. MyTM fails to preserve the background (missing dark shadow in the lower-left region) and misses the accessory (earrings). FaceTT maintains consistent background and accessory details.
    % and achieves natural and accurate aging transformations, particularly at older ages. 
    }
    \label{fig:comparison_mytm}
    \vspace{-1em}
\end{figure}

Table~\ref{tab:table5} depicts a comparison of the FFHQ-Aging dataset where the proposed FaceTT achieves a better MAE score, indicating improved aging accuracy, with a 15\% improvement over FADING. Furthermore, the proposed method obtains better gender accuracy across all age brackets, highlighting its ability to preserve essential identity attributes as age progresses. The KID analysis further supports our results. The HRFAE exhibits minimal changes across ages (see Figure~\ref{fig:visual-1}), resulting in a low KID score.

Figure~\ref{fig:visual-1} qualitatively validates that the proposed FaceTT achieves better visual fidelity with minimal artifacts. It generates an image in ~5 seconds, while FADING takes around 130 seconds. This demonstrates that our framework enhances reconstruction quality with a significant reduced inference time. A detailed analysis with more examples is provided in \texttt{supplementary-S2}.

\noindent \textbf{Comparison with Recent Unpublished Works:}
% Additionally, we provide visual comparisons with two recent unpublished works—Aging Multiverse\footnote{\url{https://agingmultiverse.github.io/}}~\cite{gong2025agingmultiverse} and MyTimeMachine (MyTM)\footnote{\url{https://mytimemachine.github.io/}}~\cite{qi2025mytimemachine}—using the results available on their official webpages.
% Figure~\ref{fig:comparison_aging_multiverse} shows results across two target age groups. Compared to Aging Multiverse, our method produces more realistic age progression while better preserving external factors, background details and skin tone. Notably, our model accurately retains accessories such as eyeglasses, which are not reproduced by Aging Multiverse, and it maintains background consistency, where Aging Multiverse introduces noticeable scene alterations.
% When compared to MyTM work (Figure~\ref{fig:comparison_mytm}), Which emphasizes personalized temporal aging, our method achieves natural and accurate aging transformations, particularly at older ages, while also effectively preserving background details and object consistency such as earrings. This improvement also aligns with our higher $\text{ID}_{sim}^{ref}$ scores, further confirming the superior identity preservation of our method.
\noindent Additionally, we include visual comparisons with two recent unpublished works—Aging Multiverse\footnote{{agingmultiverse.github.io/}}~\cite{gong2025agingmultiverse} and MyTimeMachine (MyTM)\footnote{{mytimemachine.github.io/}}~\cite{qi2025mytimemachine}—using
the results available on their official webpages. As shown in Figure~\ref{fig:comparison_aging_multiverse}, our method produces more realistic attribute-aware age progression than Aging Multiverse, while preserving skin tone, background details, and accessories (e.g., eyeglasses). When compared to MyTM (Figure~\ref{fig:comparison_mytm}), which emphasizes personalized temporal aging, our method achieves natural and accurate aging transformations, particularly at older ages, while also effectively preserving background details and object consistency such as earrings. These findings are consistent with our improved $\text{ID}_{sim}^{ref}$ scores, highlighting superior identity preservation.

\begin{table}[t!]
\centering
\caption{Comparison of $\text{ID}\_{sim}$ scores across different methods.}
\label{tab:id_sim_comparison}
\vspace{-1em}
\begin{adjustbox}{width=0.42\textwidth}
\begin{tabular}{@{}lccc>{\columncolor{lightmintbg}}c@{}}
\toprule
$\text{ID}\_{sim}$ & \textbf{HRFAE} & \textbf{CUSP} & \textbf{FADING} & \textbf{FaceTT} \\ \midrule
$\text{ID}_{sim}^{cyc}$ in FFHQ test-set & 0.61 & 0.55 & 0.67 & 0.69 \\
$\text{ID}_{sim}^{cyc}$ in Celebrity Test-set & 0.60 & 0.39 & 0.77 & 0.80 \\
$\text{ID}_{sim}^{ref}$ in Celebrity Test-set & 0.42 & 0.32 & 0.50 & 0.55 \\ 
\bottomrule
\end{tabular}
\end{adjustbox}
\vspace{-10pt}
\end{table}
\begin{figure}[t!]
    \centering 
    \includegraphics[width=0.475\textwidth]{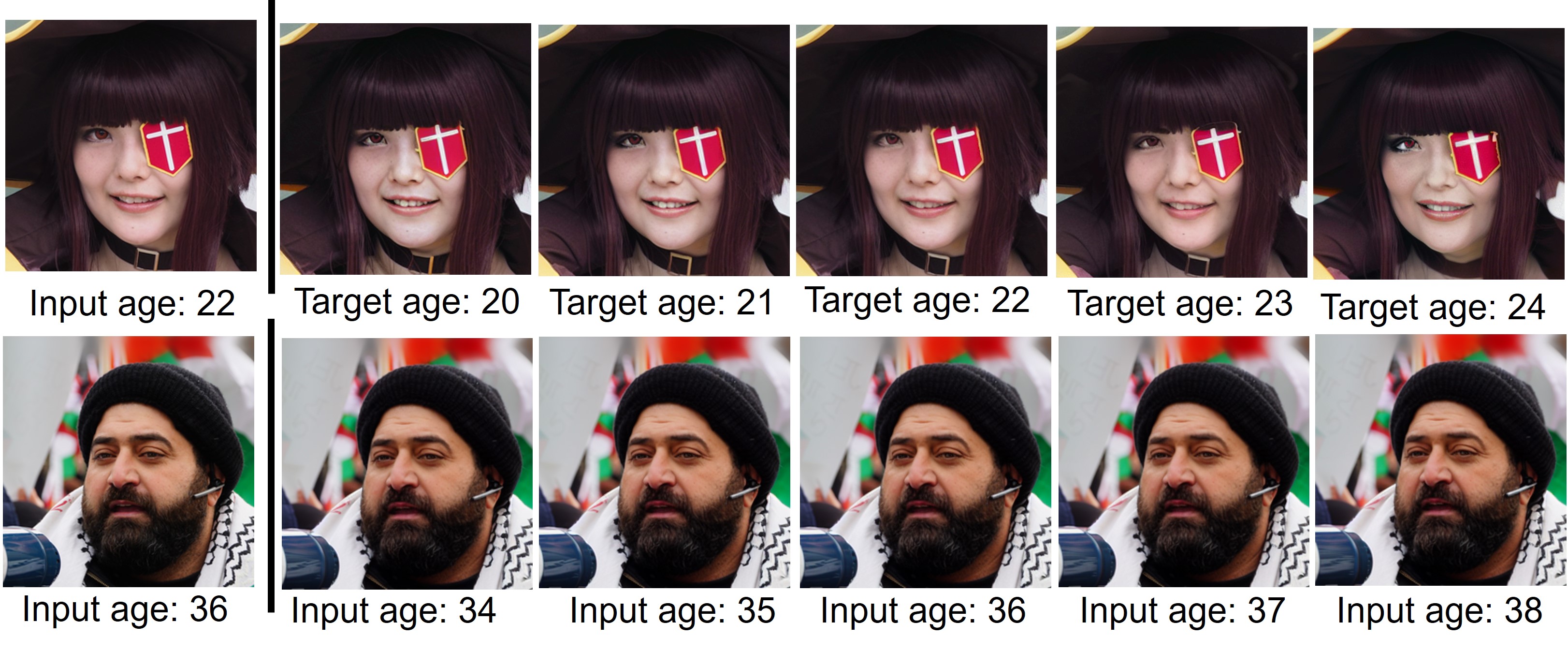}
    \vspace{-2.25em}
    \caption{Short-Range Aging analysis: FaceTT maintains strong identity consistency across closely spaced age targets. 
    % ensuring that fine-grained age transformations do not alter facial geometry or personal characteristics.
    }
    \label{fig:identity_preservation}
    \vspace{-0.7em}
\end{figure}
\begin{table}[t!]
\centering
% \vspace{-0.5em}
\caption{Biometric matching results between original and re-aged images on FFHQ dataset. The metrics are False Non Match Rate (FNMR) at False Match Rate (FMR) = 0.01/0.1\%. Lower is better.}
\label{tab:biometric-comparison-table}
\vspace{-1em}
\begin{adjustbox}{width=0.475\textwidth}
\begin{tabular}{ccccccc}
\hline
\textbf{Method} & \textbf{2} & \textbf{12} & \textbf{25} & \textbf{35} & \textbf{45} & \textbf{60} \\
\hline
CUSP & 0.52/0.28 & 0.22/0.11 & 0.31/0.13 & 0.27/0.09 & 0.23/0.07 & 0.45/0.10 \\

FADING & 0.55/0.49 & 0.11/0.07 & 0.26/0.09 & 0.25/0.07 & 0.25/0.07 & 0.33/0.12 \\

\rowcolor{lightmintbg}
FaceTT & 0.60/0.49 & 0.18/0.10 & 0.06/0.06 & 0.02/0.01 & 0.03/0.02 & 0.13/0.07 \\
\hline
\end{tabular}
\end{adjustbox}
\vspace{-1em}
\end{table}
\begin{figure*}[t!]
    \centering 
    \includegraphics[width=\textwidth]{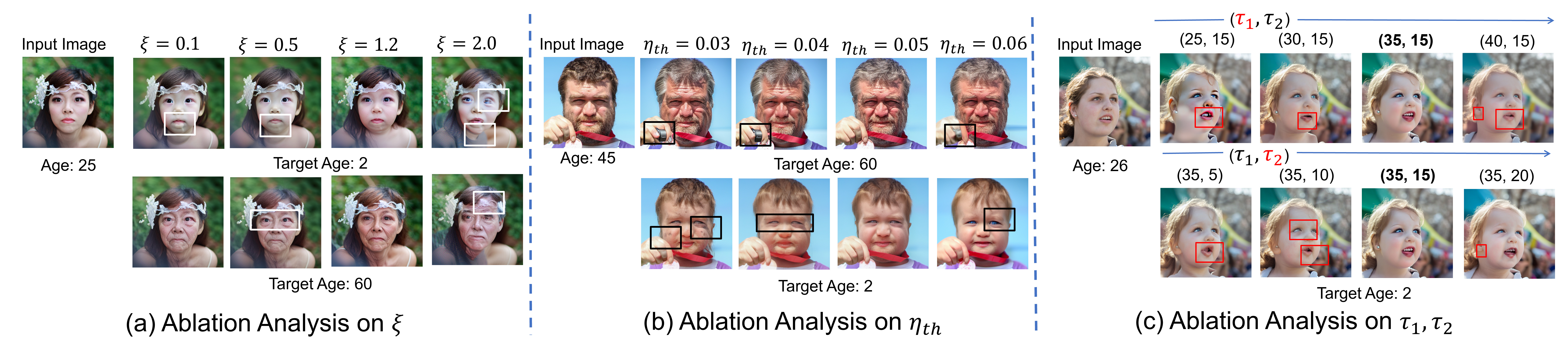}
    \vspace{-2.5em}
    \caption{Ablation analysis on hyper-parameters (a) $\xi$; (b) $\eta_{th}$; (c) $\tau_{1}$, $\tau_{2}$. (More examples are provided in \texttt{supplementary-S3}).}
    \label{fig:ablation_hyper-parameters}
    \vspace{-1.2em}
\end{figure*}
\begin{figure}[t!]
    \centering 
    \vspace{-0.5em}
    \includegraphics[width=0.475\textwidth]{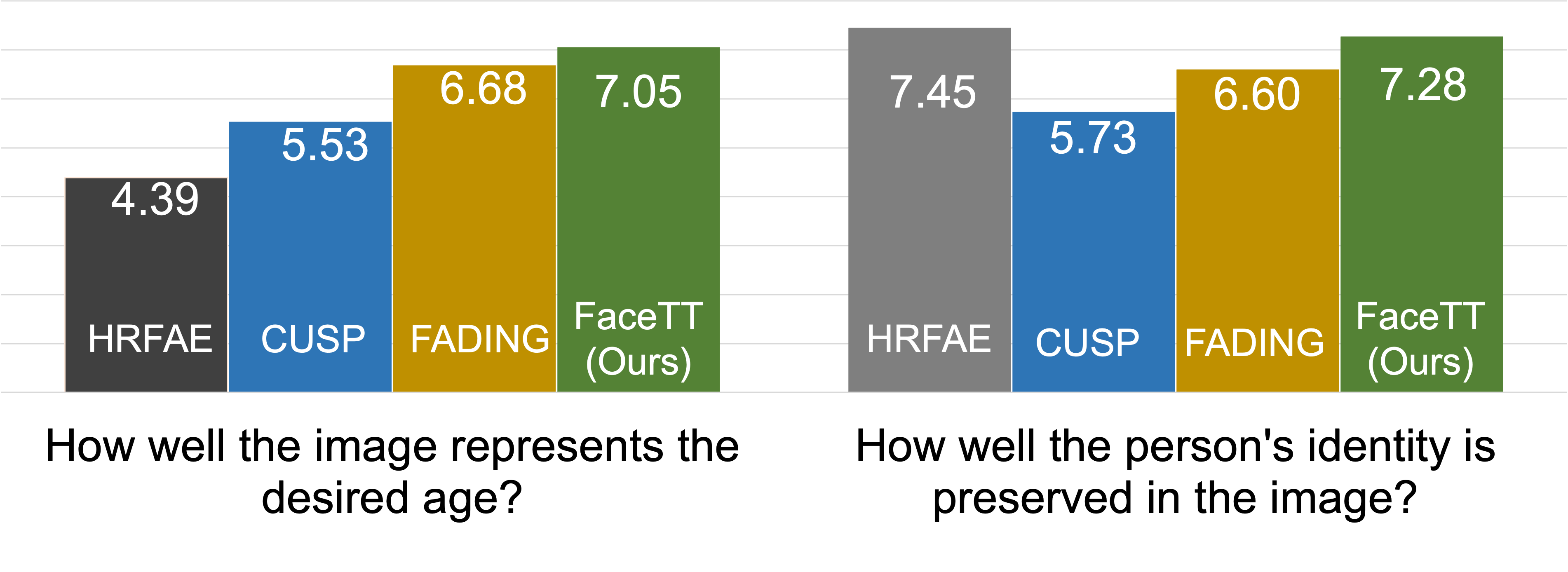}
    \vspace{-25pt}
    \caption{User study Analysis: our method consistently scores high in both aging and identity preservation. HRFAE's high score for identity preservation stems from its limited aging ability, often producing results nearly identical to the input.}
    \label{fig:user_study}
    \vspace{-5pt}
\end{figure}
\begin{table}[t!]
\centering
% \vspace{-0.5em}
\caption{Impact of proposed components on CelebA-HQ for young-to-60 task.}
\label{tab:ablation-inversion-editing}
\vspace{-1em}
\begin{adjustbox}{width=0.49\textwidth}
\begin{tabular}{@{}cccccccc@{}}
\toprule
\textbf{Angular Inversion} & \textbf{AAC} & \textbf{Predicted Age} & \textbf{Blur $\downarrow$} & \textbf{Gender $\uparrow$} & \textbf{Smiling $\uparrow$} & \textbf{Neutral $\uparrow$} & \textbf{Happy $\uparrow$} \\ \midrule
\multicolumn{2}{c}{\quad \quad Real Images} & 65.14 $\pm$ 4.86 & 1.73 & --- & --- & --- & --- \\ 
\ding{55}                  & \ding{55}     &   69.88 $\pm$ 6.20                    &    2.18           &    98.44             &          76.17        &       56.54           &         73.19       \\
\ding{55}                   & \ding{51}    &     61.70 $\pm$ 6.30                  &  2.21            &     99.22            &      73.78            &       64.99           &   66.72             \\
\ding{51}              & \ding{55}     &        61.25 $\pm$     6.20           &    2.89          &    99.02                &     68.58            &      61.37            & 64.48               \\
\rowcolor{lightmintbg} \ding{51}              & \ding{51}    &     62.05 $\pm$   6.81                 &         2.18      &      99.79           &       78.31           &     62.67             &      72.17          \\ \bottomrule
\end{tabular}
\end{adjustbox}
\label{tab:angular_aac}
% \vspace{-5pt}
\end{table}
\begin{figure}[t!]
    \centering
    \vspace{-0.5em}
    \includegraphics[width=0.48\textwidth, height=4.5cm]{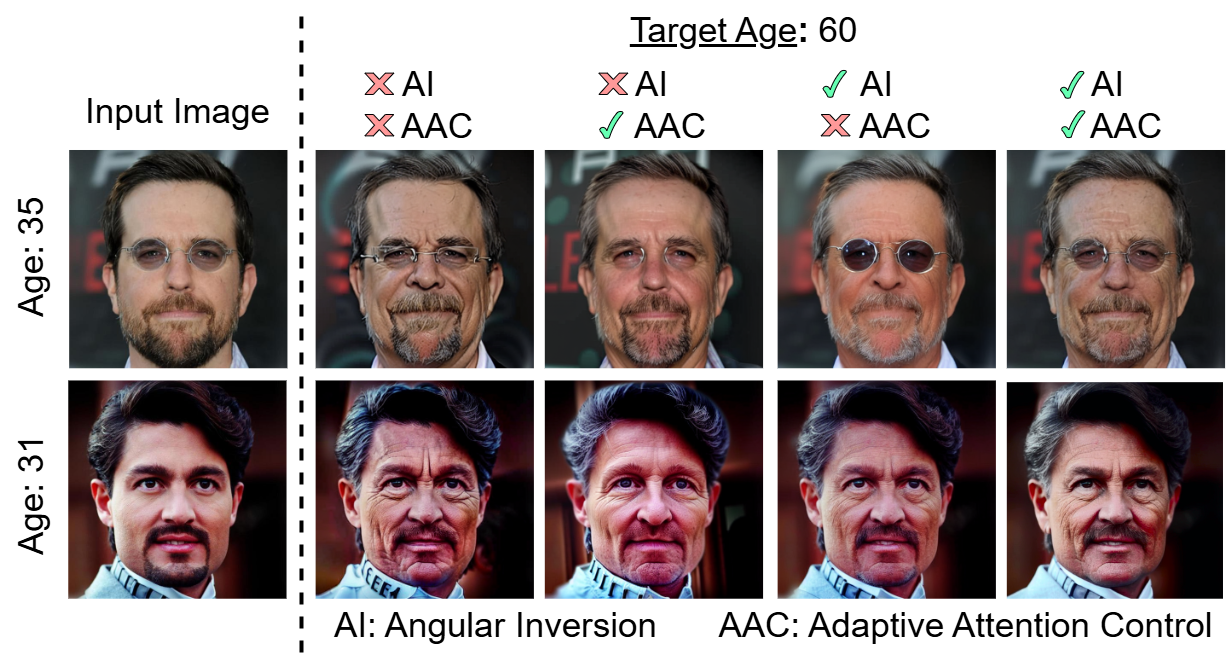}
    \vspace{-2em}
    \caption{\footnotesize Ablation analysis on CelebA-HQ for young-to-60 task.}
    \label{fig:ablation}
    \vspace{-1em}
\end{figure}

\subsection{Evaluation on Identity Preservation}
\noindent \textbf{Cyclic / Reference based Aging Analysis:} Table~\ref{tab:id_sim_comparison} reports the $\text{ID}_{\text{sim}}$ scores for face aging methods on both the FFHQ test set and the celebrity age progression test set. For the FFHQ dataset, we divide samples into \emph{young} ($<$40) and \emph{old} ($>$40) groups. Faces in the young group are aged to 60 and then reverted back to compute the $\text{ID}_{\text{sim}}^{\text{cyc}}$, while faces in the old group are de-aged to 20 and then restored to compute the same measure. This bidirectional procedure provides a balanced evaluation of both aging and de-aging identity consistency.
For the celebrity test set, we report both $\text{ID}_{\text{sim}}^{\text{cyc}}$ and $\text{ID}_{\text{sim}}^{\text{ref}}$. The cyclic score is computed by averaging $\text{ID}_{\text{sim}}^{\text{cyc}}$ across all possible age-range cycles.
% (20$\rightarrow$40$\rightarrow$20, 20$\rightarrow$60$\rightarrow$20, 40$\rightarrow$20$\rightarrow$40, 40$\rightarrow$60$\rightarrow$40, 60$\rightarrow$20$\rightarrow$60, and 60$\rightarrow$40$\rightarrow$60). 
The $\text{ID}_{\text{sim}}^{\text{ref}}$ compares each re-aged image to the corresponding real target-age image.
Across both datasets, the proposed FaceTT consistently achieves the highest $\text{ID}_{\text{sim}}^{\text{cyc}}$ and $\text{ID}_{\text{sim}}^{\text{ref}}$ scores, indicating superior identity preservation and more stable bidirectional age transformations compared to others.

\noindent \textbf{Short-Range Aging Analysis:}
% While the proposed $\text{ID}_{\text{sim}}^{\text{cyc}}$ \& $\text{ID}_{\text{sim}}^{\text{ref}}$ quantitatively assess identity consistency, 
We further conduct a qualitative analysis to assess identity retention under fine-grained age changes. Figure~\ref{fig:identity_preservation} shows examples where each face is re-aged within a narrow range around the original age (e.g., $x-2 \leq \text{target age} \leq x+2$). Our model produces visually consistent faces across these small age shifts, indicating strong identity stability. 
% This fine-grained evaluation complements the cyclic and reference-based metrics, confirming that the proposed framework preserves identity not only under large age transitions but also across subtle, short-range age variations.

% Here, we perform a qualitative analysis to visually verify identity preservation under fine-grained age transformations. Figure~\ref{fig:identity_preservation} illustrates examples where an individual’s image is aged within a short-range 
% around the input age (e.g., \(x - 2 \leq \text{target age} \leq x + 2\)). 
% In this setting, we age each input image to nearby target ages using our model and compare the resulting faces.
% Here we can see that, our method consistently maintains high facial similarity across all short-range transformations, demonstrating robust identity retention. 
% This short-range aging evaluation complements the cyclic and reference-based identity metrics, confirming that the proposed framework achieves not only large-scale age translation consistency but also fine-grained, identity-stable transformations across adjacent age intervals.
% If the re-aged outputs remain visually indistinguishable from the input in terms of global facial structure, local features (such as eyes, nose, and mouth geometry), and overall identity cues, it indicates that the model successfully preserves the individual’s identity even when exposed to subtle aging or de-aging perturbations. 

\noindent \textbf{Biometric Matching Analysis:}
Inspired by \cite{Idendity_biometric}, we perform a biometric verification study on the FFHQ dataset. Following the standard verification protocol, we report the \textit{False Non-Match Rate (FNMR)} at two operating points of \textit{False Match Rate (FMR)} = 0.01 and 0.1\%. This experiment measures how well aged faces remain recognizable to a pretrained face recognition model \cite{deng2019arcface}. Table~\ref{tab:biometric-comparison-table} presents FNMR@FMR across six aging distances (2, 12, 25, 35, 45, and 60 years) for CUSP, FADING, and our FaceTT. Lower FNMR indicates stronger identity retention. FaceTT consistently achieves the lowest FNMR (particularly for the middle and older age groups). Notably, at a 35-year aging distance, FaceTT achieves an FNMR of 0.02/0.01, representing a significant improvement over competing methods.

% To further validate the identity-preserving capability of our model under re-aging transformations, 
% we conduct a biometric evaluation on the FFHQ dataset using a face verification protocol. 
% Specifically, we measure the False Non-Match Rate (FNMR) at two operating points of 
% False Match Rate (FMR) = 0.01 and 0.1\%, following standard biometric verification practices. 
% This experiment assesses how well re-aged faces can be matched with their original identities 
% using a pre-trained face recognition model.

% Table~\ref{tab:biometric-comparison-table} reports FNMR@FMR for multiple re-aging levels, 
% where lower FNMR indicates better identity retention. 
% We compare three methods—CUSP, FADING, and our FaceTT—across six re-aging distances 
% (2, 12, 25, 35, 45, and 60 years). 
% As the re-aging gap increases, the FNMR naturally rises for all methods, reflecting greater visual drift from the original identity. 
% However, our FaceTT achieves consistently lower FNMR values across all ranges, 
% demonstrating superior robustness in preserving identity even under large age transitions (up to 60 years). 
% In particular, FaceTT attains an FNMR of 0.02/0.01 at 35 years of re-aging, 
% showing a 2–3$\times$ reduction compared to competing methods.

% These results, together with our cyclic and reference-based evaluations, 
% establish that FaceTT not only maintains identity fidelity under small and cyclic transformations 
% but also achieves state-of-the-art biometric consistency across extensive age variations.

\noindent \textbf{User Study Analysis:}
% We conducted a user study with 35 anonymous participants using 14 images (10 from FFHQ and 4 in-the-wild), re-aged to six target ages (2, 12, 25, 35, 45 and 60). Participants rated each image (1–10) based on how well it represented the target age and preserved identity. The mean opinion score, shown in Figure~\ref{fig:user_study} (a), indicates our framework received the highest rating in achieving the target age and 2nd highest rating in identity preservation. HRFAE's high score for identity preservation stems from its limited re-aging ability, often producing results nearly identical to the input (see Figure~\ref{fig:visual-1}). 
We conducted a user study with 35 participants on 14 images (10 from FFHQ and 4 in-the-wild) re-aged to six target ages (2, 12, 25, 35, 45 and 60). Participants rated each result (1–10) for age accuracy and identity preservation. As shown in Figure~\ref{fig:user_study}, our method ranks highest in achieving the target age and second-highest in identity preservation. HRFAE scores high in identity preservation mainly because it performs minimal aging and often produces images nearly identical (Figure~\ref{fig:visual-1}).

\subsection{Ablation Analysis}
\noindent \textbf{Impact of individual components:}
To validate the impact of our proposed \textit{Angular Inversion} and \textit{AAC} modules, we conducted ablation studies on the CelebA-HQ dataset. The results, presented in Table~\ref{tab:angular_aac} and Figure~\ref{fig:ablation} demonstrate that both the \textit{Angular Inversion} and \textit{AAC} modules contribute to improving the overall performance.

\noindent \textbf{Sensitivity Analysis of Hyper-parameters:}  
Figure~\ref{fig:ablation_hyper-parameters} presents the ablation analysis of $\xi$, $\eta_{th}$, $\tau_1$, and $\tau_2$.
\begin{itemize}
    \item In Figure~\ref{fig:ablation_hyper-parameters}(a), low $\xi$ (0.1, 0.5) yields minimal change—identity is preserved but aging is weak—while high $\xi$ (2.0) causes excessive edits and identity loss. The optimal balance is achieved at $\xi = 1.2$.
    \item Figure~\ref{fig:ablation_hyper-parameters}(b) shows that a low $\eta_{th}$ (0.03) over-preserves source traits, limiting aging, whereas a high value (0.06) over-modifies details and introduces artifacts. $\eta_{th}=0.05$ produces the most natural aging with consistent identity.
    \item In Figure~\ref{fig:ablation_hyper-parameters}(c), low $\tau_1$ (25) retains too many source attributes, while high values (40) cause exaggerated changes. $\tau_2$ similarly controls transformation smoothness. The optimal configuration $(\tau_1,\tau_2) = (35,15)$ yields realistic aging with strong structural consistency.
\end{itemize}

\section{Conclusion}
% This paper introduces FaceTT, a novel framework for face aging that addresses key limitations of existing approaches. Here, we introduce \textit{Face-Attribute-Aware Prompt Refinement} that encodes both biological and environmental aging cues to enable context-aware conditioning within diffusion models. Our \textit{Angular Inversion} technique efficiently maps real images into the latent space of a pre-trained diffusion model, ensuring high-fidelity reconstruction without iterative optimization. The \textit{Adaptive Attention Control} (\textit{AAC}) mechanism dynamically balances cross-attention for semantic age-specific edits and self-attention for structural preservation, achieving precise and identity-consistent transformations. In future work, we plan to extend FaceTT for temporal age progression in videos, incorporating motion and expression consistency across frames.

In this paper, we present FaceTT, a novel diffusion-based framework for realistic and identity-consistent face aging, enabled by Face-Attribute-Aware Prompt Refinement, Angular Inversion, and Adaptive Attention Control. Our experiments demonstrate strong aging realism, background preservation, and superior identity retention compared to existing methods. In future work, we plan to extend FaceTT to support temporal age progression in videos, incorporating motion and expression consistency across frames, thereby broadening its applicability to real-world scenarios.

%% file: sec/X_suppl.tex
\clearpage
\setcounter{page}{1}
\renewcommand{\thesection}{S\arabic{section}}
\renewcommand{\thetable}{S\arabic{table}}
\renewcommand{\thefigure}{S\arabic{figure}}
\setcounter{section}{0}
\setcounter{figure}{0}
\setcounter{table}{0}

% \maketitlesupplementary

\twocolumn[{
    \renewcommand\twocolumn[1][]{#1} 
    % \vspace{-8mm}
    \maketitlesupplementary
    \vspace{-6mm}
    \begin{center}
        \centering
        \includegraphics[width=0.87\linewidth]{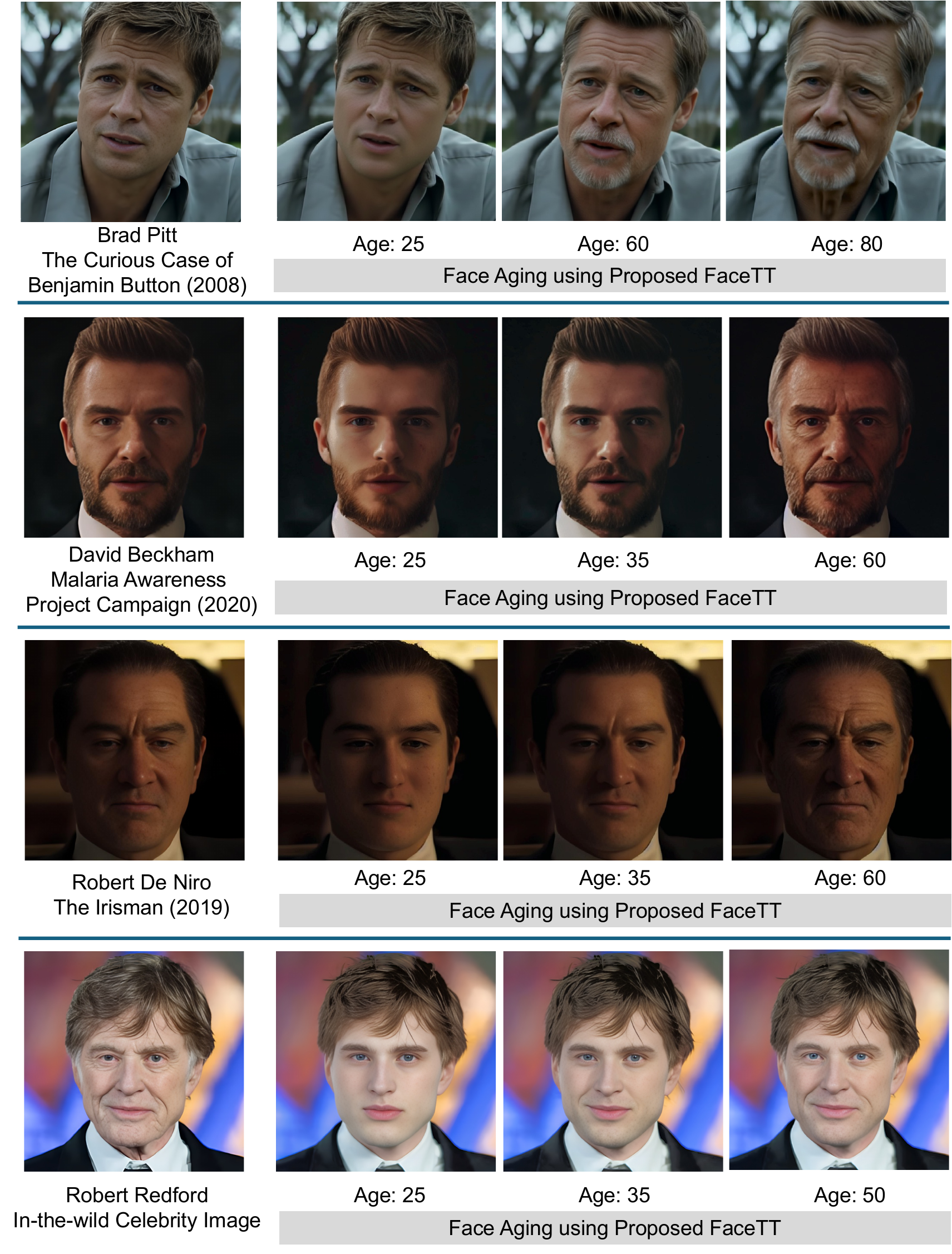}
        \vspace{-2mm}
        \captionof{figure}{Results of our proposed method FaceTT in Film/Media application. FaceTT produces realistic and identity-preserving age progression on movie and in-the-wild celebrity images. These results demonstrate visually coherent transitions across young, middle-aged, and elderly appearances.}
    \label{fig:teaser}
    \end{center}
}]

This supplementary presents below contents which we could not include in the main paper due to space constraints:
% \tableofcontents
%%%%%%%%%%%%%%%%%%%%%%%%%%%%%%%%% Table of Contents %%%%%%%%%%%%%%%
\section*{Table of Contents}
\addcontentsline{toc}{section}{Table of Contents}
\begin{enumerate}
    \item[S1] \hyperref[supp:analysis-attention]{Analysis on Cross- and Self-Attention} \dotfill \pageref{supp:analysis-attention}
    \begin{enumerate}
        \item[S1.1] \hyperref[supp:probing]{Probing Analysis} \dotfill \pageref{supp:probing} 
    \end{enumerate}
    \item[S2] \hyperref[supp:additional-visual]{Additional Visual Comparison} \dotfill \pageref{supp:additional-visual}
    % \begin{enumerate}
    %     \item[S2.1] \hyperref[supp:face-reaging-methods-comparison]{Analysis on Face Re-aging methods} \dotfill \pageref{supp:face-reaging-methods-comparison} 
    %     \item[S2.2] \hyperref[supp:unpublished-work-analysis]{ Analysis on Unpublished work} \dotfill \pageref{supp:unpublished-work-analysis} 
    % \end{enumerate}
    \item[S3] \hyperref[supp:analysis-identity]{Analysis on Identity Preservation} \dotfill \pageref{supp:analysis-identity}
    \begin{enumerate}
        \item[S3.1] \hyperref[supp:cyclic-aging-analysis]{Cyclic / Reference based Aging Analysis} \dotfill \pageref{supp:cyclic-aging-analysis} 
        \item[S3.2] \hyperref[supp:Short-Range-Aging-Analysis]{ Short-Range Aging Analysis} \dotfill \pageref{supp:Short-Range-Aging-Analysis} 
    \end{enumerate}
    \item[S4] \hyperref[supp:ablation]{Additional Ablation Analysis} \dotfill \pageref{supp:ablation}
    \begin{enumerate}
        \item[S4.1] \hyperref[supp:visual-hyper-parameter]{Visual Analysis on Hyper-Parameters} \dotfill \pageref{supp:visual-hyper-parameter} 
        \item[S4.2] \hyperref[supp:statistical-hyper-parameter]{Statistical Analysis on Hyper-Parameters} \dotfill \pageref{supp:statistical-hyper-parameter} 
    \end{enumerate}
    
    %\item[S6] \hyperref[supp:Age-Progression]{Age-Progression Analysis on Ethnicities} \dotfill \pageref{supp:Age-Progression}
    \item[S5] \hyperref[supp:visual-inversion-editing]{Analysis on Inversion \& Editing Techniques} \dotfill \pageref{supp:visual-inversion-editing} 
    \item[S6] \hyperref[supp:ethical]{Ethical Concerns} \dotfill \pageref{supp:ethical}
    \item[S7] \hyperref[supp:baselines]{Baseline Details} \dotfill \pageref{supp:baselines}
\end{enumerate}

\section{ Analysis on Cross- and Self-Attention} \label{supp:analysis-attention}
We investigate how cross-attention and self-attention in Stable Diffusion influence text-guided image editing through the U-Net model's core components. Following the standard from \cite{p2p}, spatial features are linearly projected to form queries ($Q$), while text features are transformed into keys ($K$) and values ($V$) for the cross-attention module. In the self-attention module, $K$ and $V$ are derived from spatial features. The attention mechanism is defined as:
$\text{Attention}(K, Q, V) = MV = \text{softmax} \left( \frac{QK^T}{\sqrt{d}} \right) V$,
where $M_{i,j}$ represents the attention weights for aggregating the $j$-th token's value at pixel $i$, with $d$ being the dimensionality of $K$ and $Q$.

Semantic modifications from natural language are categorized as rigid or non-rigid. Rigid changes, like background or visual element alterations, are handled using cross-attention \cite{p2p}, while non-rigid changes, such as adding/removing objects or changing actions, are managed with self-attention \cite{FPE}. 
To evaluate whether the cross- and self-attention maps in face images contain meaningful semantic information, we employ a probing approach inspired by natural language processing (NLP) techniques \cite{prob1, prob2}, as demonstrated in \cite{FPE}.

\begin{figure}[t]
    \centering
    \includegraphics[width=0.5\textwidth]{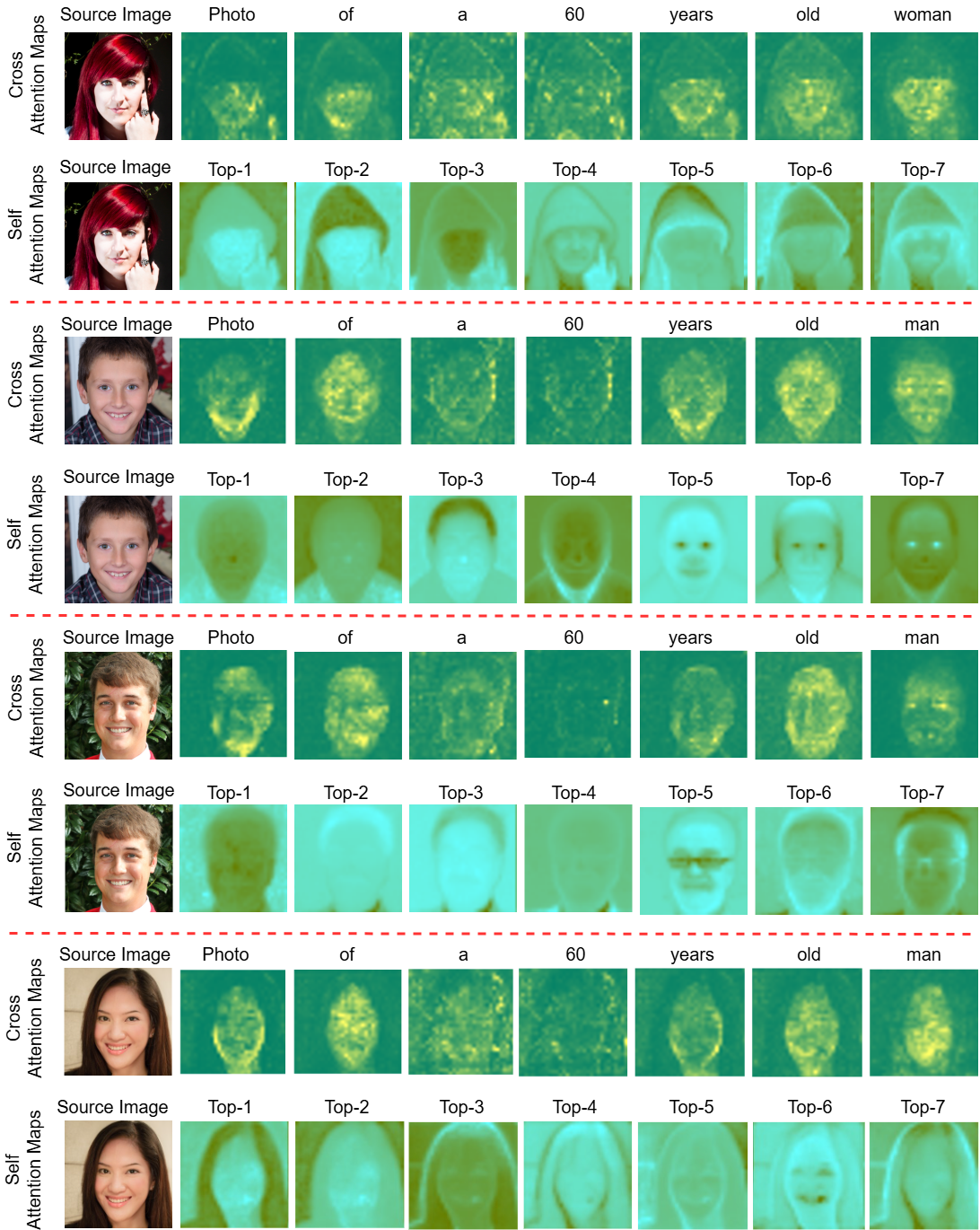}
    \vspace{-1.5em}
    \caption{Visualization of Cross- \& Self-attention maps for age progression to 60-year-old appearance. Cross-attention maps highlight regions in the source image (e.g., eyes, nose, mouth) that align with target prompt \texttt{<Photo of a 60 years old woman/man>}. While self-attention visualization showcases top-7 components derived from singular value decomposition (SVD) \cite{SVD}, which is important to maintain identity and consistency.}
    \label{fig:visual-attention-supp}
    \vspace{-1em}
\end{figure}
\vspace{-0.5em}
\subsection{ Probing Analysis:}
\label{supp:probing}
Following \cite{FPE}, we design and train a task-specific classifier consisting of a two-layer MLP to predict semantic categories from the attention maps. Using the prompt \texttt{Photo of a <2/10/30/50/70> years old <man/woman/boy/girl>}, we examine the effectiveness of the attention mechanisms. In this probing analysis, we avoid using our Face-Attribute-Aware refined prompts because they introduce additional semantic cues (e.g., skin texture, lifestyle conditions) that go beyond age and gender. Including these details would make the probe rely on prompt semantics rather than the intrinsic information encoded in the attention maps. Using simple age-and-gender prompts ensures a controlled setting where we can directly evaluate what the model’s native attention layers capture without external influence.

\begin{table}[t!]
    % \vspace{-1em}
    \caption{Probing accuracy of cross-attention maps w.r.t. different tokens. L denotes layers of the U-Net model.} \label{table:prob-cross-attention}
    \vspace{-1em}
    \centering
    \footnotesize
    \begin{adjustbox}{width=0.46\textwidth}
    \begin{tabularx}{\columnwidth}{lcccccccc}
        \toprule
        Class & L-3 & L-6 & L-9 & L-10 & L-12 & L-14 & L-16 & Avg. \\
        \midrule
        Age 2 & 0.29 & 0.42 & 0.69 & 0.81 & 0.83 & 0.85 & 0.75 & 0.77 \\
        Age 10 & 0.60 & 0.20 & 0.70 & 0.92 & 0.90 & 0.90 & 0.63 & 0.69 \\
        Age 30 & 0.57 & 0.92 & 0.80 & 0.88 & 0.89 & 0.89 & 0.78 & 0.79 \\
        Age 50 & 0.50 & 0.87 & 0.80 & 0.72 & 0.87 & 0.87 & 0.41 & 0.81 \\
        Age 70 & 0.54 & 0.41 & 0.82 & 0.71 & 0.63 & 0.55 & 0.52 & 0.59 \\
        \midrule
        man & 0.67 & 0.77 & 0.86 & 0.83 & 0.81 & 0.81 & 0.90 & 0.80 \\
        woman & 0.62 & 0.80 & 0.83 & 0.78 & 0.76 & 0.82 & 0.84 & 0.78 \\
        boy & 0.60 & 0.80 & 0.90 & 0.92 & 0.96 & 0.75 & 1.0 & 0.85 \\
        girl & 0.50 & 0.70 & 0.92 & 0.84 & 0.87 & 0.70 & 0.90 & 0.78 \\
        \bottomrule
    \end{tabularx}
    \end{adjustbox}
    \vspace{-0.75em}
\end{table}
\begin{table}[t!]
    \caption{Probing analysis of cross-attention maps w.r.t. difference tokens. The upper part shows the classification accuracy corresponding to the token \texttt{Photo}, and the lower shows results for \texttt{old}. L denotes layers of the U-Net model.}    \label{table:prob-other-tokens}
    \vspace{-1em}
    \centering
    \footnotesize
    \begin{adjustbox}{width=0.46\textwidth}
    \begin{tabularx}{\columnwidth}{lcccccccc}
        \toprule
        Class & L-3 & L-6 & L-9 & L-10 & L-12 & L-14 & L-16 & Avg. \\
        \midrule
        Age 2 & 0.85 & 0.78 & 0.89 & 0.80 & 0.93 & 0.85 & 0.91 & 0.87 \\
        Age 10 & 0.80 & 0.71 & 0.80 & 0.93 & 0.50 & 0.70 & 0.50 & 0.71 \\
        Age 30 & 0.76 & 0.85 & 0.86 & 0.90 & 0.75 & 0.96 & 0.71 & 0.83 \\
        Age 50 & 0.71 & 0.73 & 0.70 & 0.83 & 0.70 & 0.55 & 0.66 & 0.70 \\
        Age 70 & 0.61 & 0.67 & 0.62 & 0.70 & 0.66 & 0.50 & 0.60 & 0.62 \\
        \midrule
        Age 2 & 0.90 & 0.83 & 0.80 & 0.92 & 0.84 & 0.92 & 0.93 & 0.88 \\
        Age 10 & 0.85 & 0.90 & 0.87 & 0.80 & 1.0 & 0.75 & 0.70 & 0.84 \\
        Age 30 & 0.80 & 0.82 & 0.79 & 1.00 & 0.70 & 0.69 & 0.75 & 0.79 \\
        Age 50 & 0.83 & 0.80 & 0.70 & 0.79 & 0.63 & 0.62 & 0.61 & 0.71 \\
        Age 70 & 0.76 & 0.71 & 0.59 & 0.83 & 0.52 & 0.70 & 0.58 & 0.67 \\
        \bottomrule
    \end{tabularx}
    \end{adjustbox}
    \vspace{-1em}
\end{table}
\begin{table}[t!]
    \caption{Probing accuracy of self-attention maps w.r.t. different tokens. L denotes layers of the U-Net model.} \label{table:prob-self-attention}
    \vspace{-1em}
    \centering
    \begin{adjustbox}{width=0.46\textwidth}
    \footnotesize
    \begin{tabularx}{\columnwidth}{lcccccccc}
        \toprule
        Class & L-3 & L-6 & L-9 & L-10 & L-12 & L-14 & L-16 & Avg. \\
        \midrule
        Age 2 & 0.00 & 0.00 & 0.00 & 0.20 & 0.00 & 0.00 & 0.00 & 0.028 \\
        Age 10 & 0.18 & 0.06 & 0.70 & 0.35 & 0.80 & 0.25 & 0.90 & 0.46 \\
        Age 30 & 0.00 & 0.00 & 0.00 & 0.00 & 0.5 & 1.0 & 0.00 & 0.21 \\
        Age 50 & 0.00 & 0.1 & 0.00 & 0.00 & 0.66 & 0.23 & 0.10 & 0.16 \\
        Age 70 & 0.00 & 0.00 & 0.82 & 0.05 & 0.25 & 0.00 & 0.23 & 0.19 \\
        \midrule
        man & 0.36 & 0.68 & 0.45 & 0.90 & 0.80 & 0.30 & 0.60 & 0.58 \\
        woman & 0.33 & 0.41 & 0.83 & 0.58 & 0.63 & 0.10 & 0.81 & 0.53 \\
        boy & 0.30 & 0.21 & 0.80 & 0.87 & 0.91 & 0.65 & 0.80 & 0.76 \\
        girl & 0.47 & 0.10 & 0.00 & 0.69 & 0.70 & 0.41 & 0.00 & 0.34 \\
        \bottomrule
    \end{tabularx}
    \end{adjustbox}
\end{table}

\textbf{Cross-Attention Maps:}
We explore the information captured by cross-attention maps by visualizing the attention patterns associated with each word in a prompt, as displayed in Figure \ref{fig:visual-attention-supp}. Each map highlights areas related to specific words, indicating that cross-attention retains rigid semantic details. In the context of the facial re-aging task, these rigid details include wrinkles and changes in skin texture, variations in the fullness of the cheeks and lips, as well as the development of age-related marks such as fine lines or sagging skin. Table \ref{table:prob-cross-attention} presents the classification performance of our trained classifier, demonstrating high accuracy, especially for the \texttt{age} and \texttt{person} categories. 

We also investigate whether cross-attention maps associated with non-edited words contain rigid information. This analysis is essential because text embeddings generated by transformer-based encoders \cite{bert, radford2021learning} preserve the contextual details of sentences. By using prompts like \texttt{Photo of a <age> years old <person>}, our experiments (shown in Table~\ref{table:prob-other-tokens}) reveal that both the tokens \texttt{Photo} and \texttt{old} exhibit significantly high classification accuracy. This finding indicates that both tokens carry rigid semantic information related to faces.

\textbf{Self-Attention Maps:}
In contrast, self-attention maps exhibit different behaviors and are analyzed similarly. Table \ref{table:prob-self-attention} shows that the classifier struggles to recognize categories (\texttt{age} and \texttt{person}) accurately, indicating that self-attention maps encode non-rigid information, or structural rather than categorical details. For our task, this structural information includes the relative positions of the eyes, nose, mouth, and the overall shape of the face (e.g., jawline). Figure \ref{fig:visual-attention-supp} illustrates that self-attention maps preserve the original structure of the image.

% \begin{figure}[t]
%     \centering
%     \includegraphics[width = 1.05\linewidth]{ICCV2025-Author-Kit/images/Attention_Visual1.pdf}
%     \vspace{-1.5em}
%     \caption{Visual Analysis of Cross- \& Self-Attention Maps. The heatmaps illustrate the cross-attention and self-attention distributions in an image generated with the prompt \texttt{Photo of a 60 years old woman}. The cross-attention heatmap depicts the contribution of each word from the prompt to the image, while the self-attention visualization showcases the top-7 components derived from singular value decomposition (SVD) \cite{SVD}.
% }
% \vspace{-0.5em}
%     \label{fig:Attention_analysis}
% \end{figure}

This show that cross-attention maps encode semantic features for category identification, while self-attention maps preserve structural integrity for image coherence.

% \begin{figure*}[t]
%     \centering
%     % \hspace{-0.5cm}
%     \includegraphics[width=\textwidth]{wacv-2026-author-kit-template/images/Fig_S2_updated (3).png}
%     \vspace{-1.3em}
%     \caption{Visual comparison of facial re-aging methods. The input images are transformed to six target ages, where our approach delivers more natural aging effects, minimal artifacts and better preservation of intricate facial details, all while requiring significantly less inference time (i.e., $\sim$ 5 sec). Zoom in for better visualization.}
%     \label{fig:visual-additional}
%     \vspace{-1em}
% \end{figure*}

% \begin{figure*}[t!]
%     \centering 
%     \includegraphics[width=0.95\textwidth]{wacv-2026-author-kit-template/images/FIg_S3_WACV.jpg}
%     % \vspace{-2.2em}
%     \caption{Visual Analysis of proposed model on different age group.}
%     \label{fig:Intermediate_age}
%     % \vspace{-1em}
% \end{figure*}

\section{Additional Visual Comparisons}
\label{supp:additional-visual}
\begin{figure*}[t]
    \centering
    % \hspace{-0.5cm}
    \includegraphics[width=\textwidth]{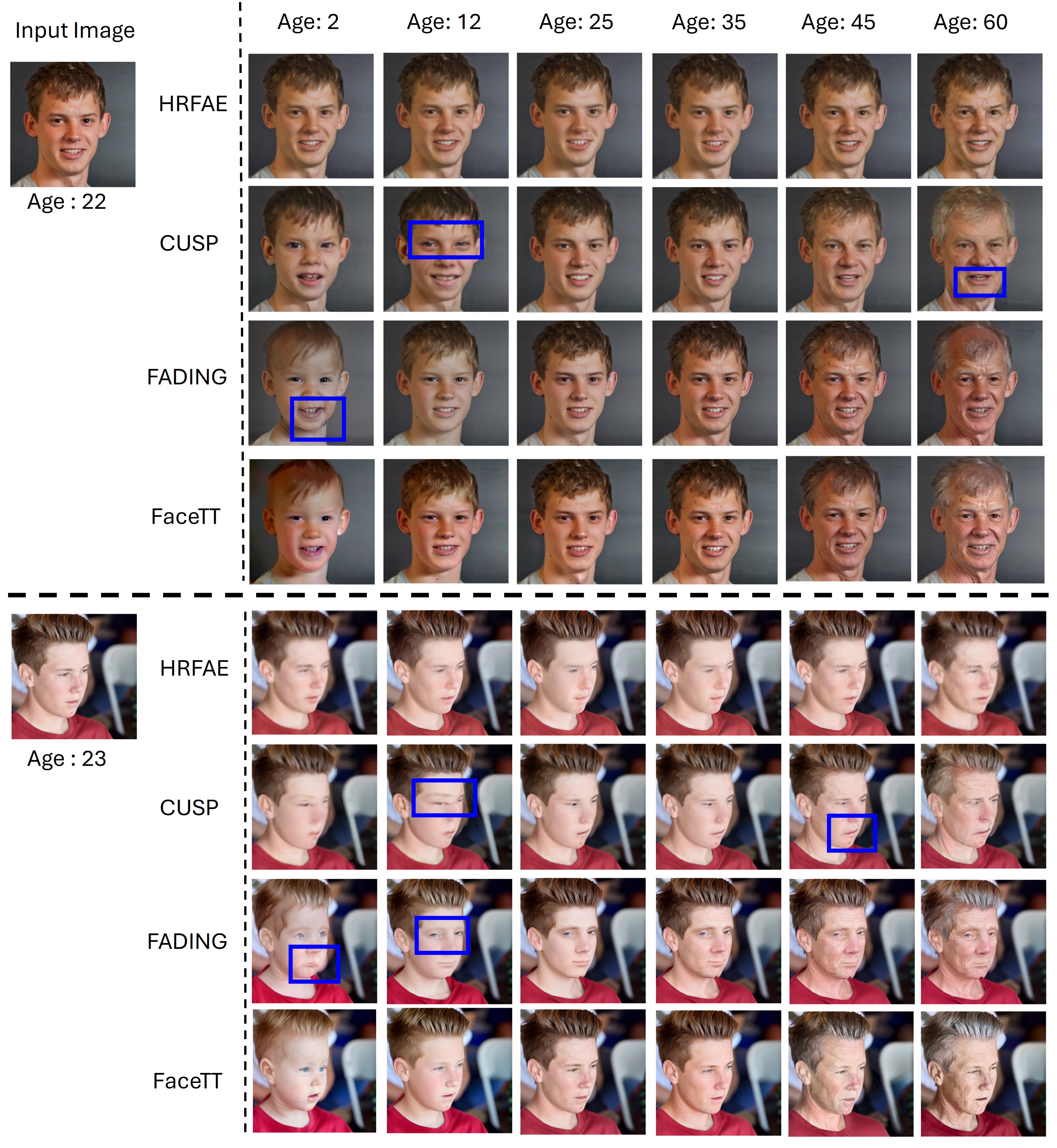}
    \vspace{-1.3em}
    \caption{Comparison of facial re-aging across different age targets using HRFAE, CUSP, FADING, and our FaceTT method (Part 1). Across both examples, FaceTT produces the most consistent identity preservation and the most realistic aging trajectories from early childhood (Age: 2) to older adulthood (Age: 60). Competing methods exhibit notable artifacts: HRFAE often fails to introduce meaningful age changes and tends to output visually similar faces across age ranges; CUSP frequently produces structural distortions and texture inconsistencies \textcolor{blue}{(blue boxes)}, especially in the eyes and mouth regions; and FADING, while capable of stronger aging effects, suffers from identity drift and unnatural facial modifications \textcolor{blue}{(blue boxes)}. In contrast, FaceTT generates smooth, coherent age transitions with preserved facial geometry and identity cues, while accurately reflecting age-specific features such as fuller cheeks in youth, subtle mid-life changes, and realistic wrinkle patterns at older ages.}
    \label{fig:visual-additional-1}
    \vspace{-1em}
\end{figure*}

\begin{figure*}[t]
    \centering
    % \hspace{-0.5cm}
    \includegraphics[width=1.03\textwidth]{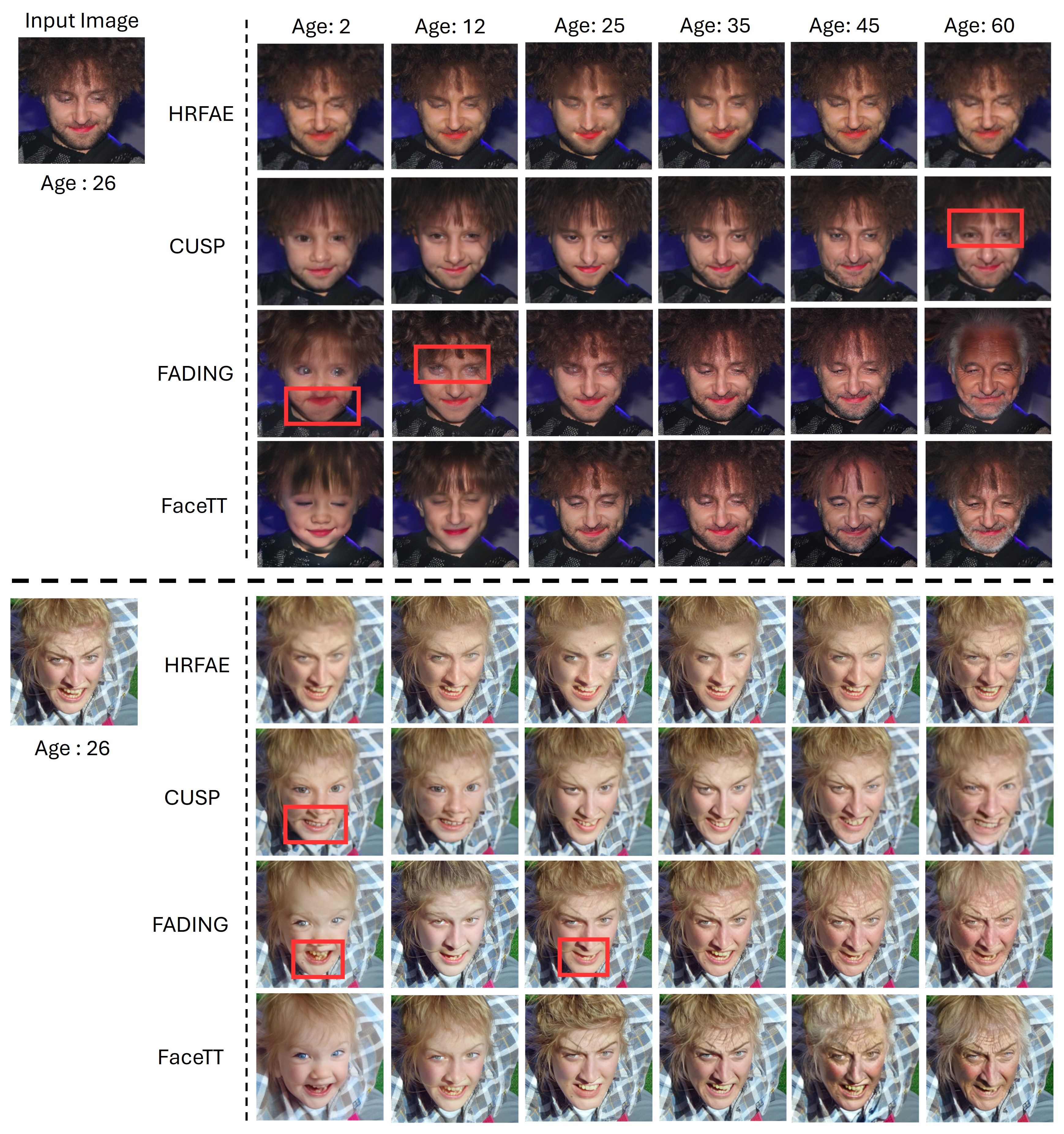}
    \vspace{-1.3em}
    \caption{Additional visual comparison of facial re-aging methods. (Part 2, continued from Figure~\ref{fig:visual-additional-1})}
    \label{fig:visual-additional-2}
    \vspace{-1em}
\end{figure*}

\begin{figure*}[t]
    \centering
    % \hspace{-0.5cm}
    \includegraphics[width=1.03\textwidth]{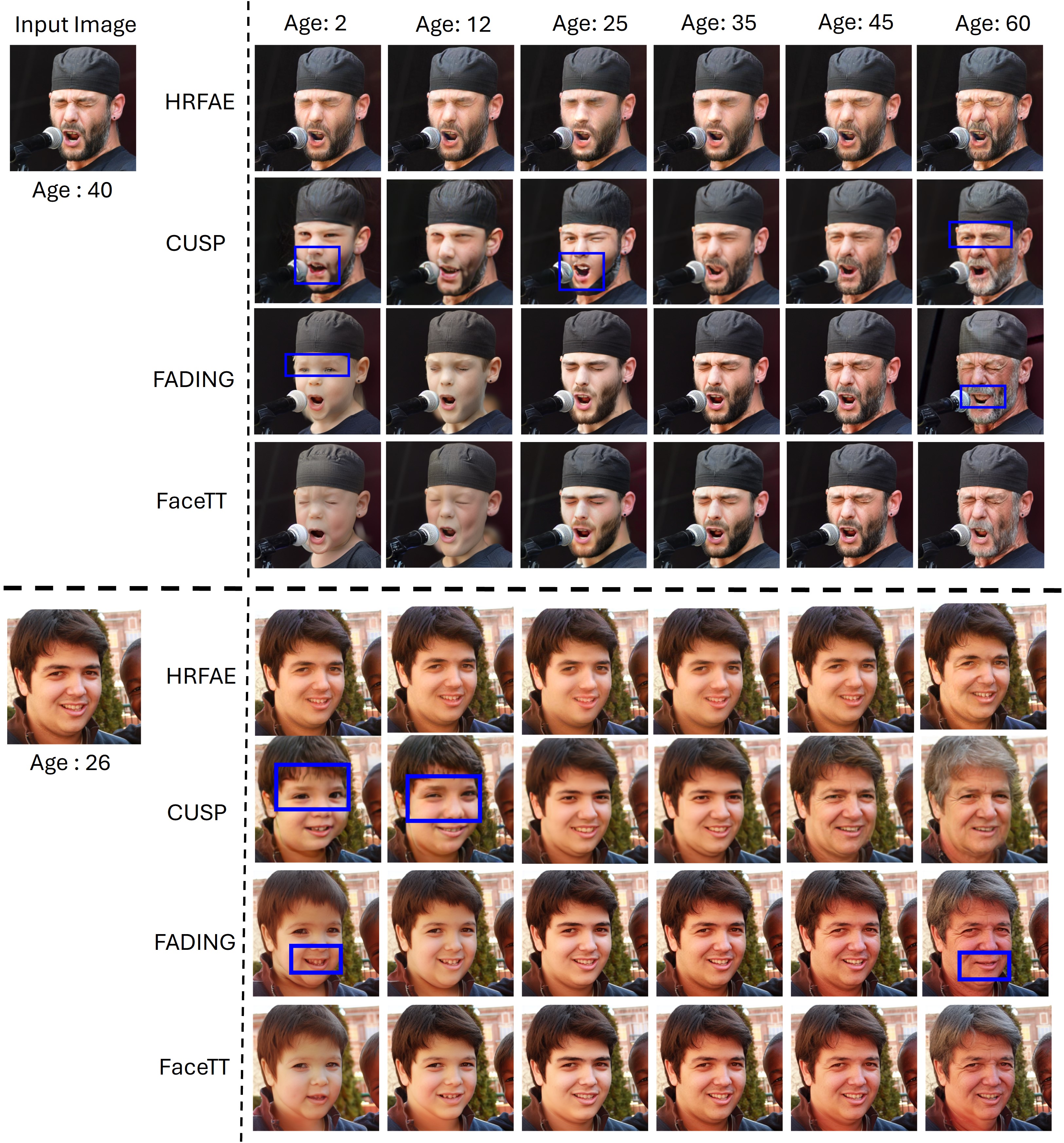}
    \vspace{-1.3em}
    \caption{Additional visual comparison of facial re-aging methods. (Part 3, continued from Figure~\ref{fig:visual-additional-1})}
    \label{fig:visual-additional-3}
    \vspace{-1em}
\end{figure*}

% \subsection{Analysis on Face Re-aging methods:}
% \label{supp:face-reaging-methods-comparison}
This section offers a visual comparison between our FaceTT and the state-of-the-art (SOTA) methods i.e., HRFAE \cite{hrfae}, CUSP \cite{cusp}, and FADING \cite{FADING}. From Figure \ref{fig:visual-additional-1} to \ref{fig:visual-additional-3} , one can observe that HRFAE preserves identity but struggles with exaggerated features. CUSP provides smooth transitions; however, it comes at the cost of losing fine details in extreme aging. FADING generates plausible results but often produces overly smooth textures or inconsistencies. In contrast, our method maintains identity and delivers highly realistic aging effects across diverse facial structures and expressions. 
% It achieves this with a faster inference speed (approximately 5 seconds compared to FADING's approximately 130 seconds), demonstrating its superior quality, efficiency, and reliability.


\section{Analysis on Identity Preservation}
\label{supp:analysis-identity}
\subsection{Cyclic / Reference based Aging Analysis:}
\label{supp:cyclic-aging-analysis}
\begin{figure*}[t]
    \centering
    \includegraphics[width=\textwidth]{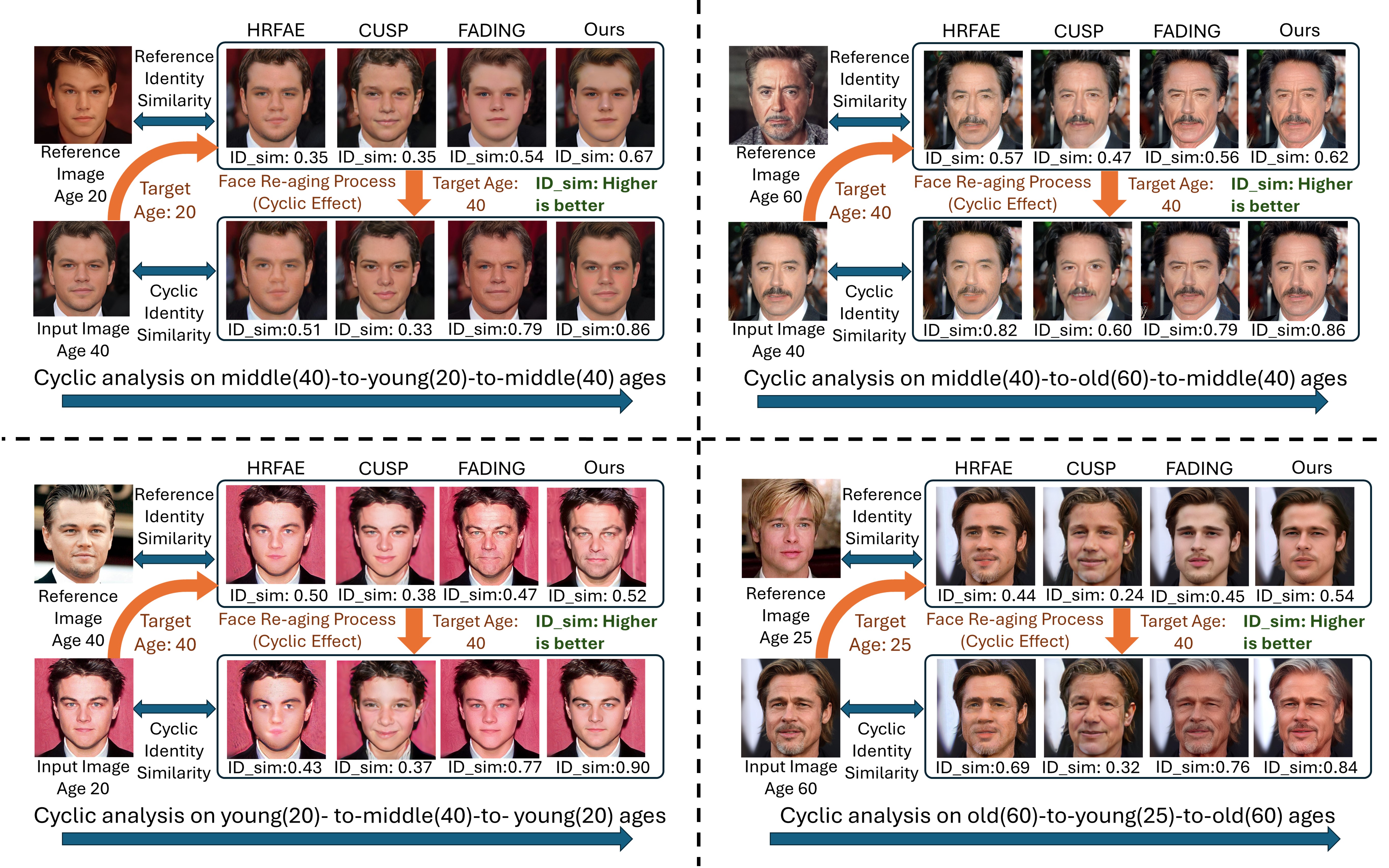}
    \vspace{-0.5em}
    % \caption{Diverse real-world use-cases of face re-aging. 
    \caption{Cyclic Identity Similarity protocol for evaluating identity preservation across age transitions. For each input celebrity image, we perform age progression and regression cycles (e.g., 40→20→40 (top left), 40→60→40 (top right), 20→40→20 (bottom left), 60→25→60 (bottom right)) and compute both Reference Identity Similarity (between the re-aged output and a real image at the target age) and Cyclic Identity Similarity (between the input and its cyclic reconstruction). Across all age cycles, our method consistently achieves higher identity similarity compared to HRFAE, CUSP, and FADING, demonstrating more faithful identity retention during both forward and reverse age transformations.}
    \label{supp:cylic-figure}
    % \vspace{-1.5em}
\end{figure*}

We conducted Cyclic and Reference-based Identity Similarity evaluation on a five-celebrity in-the-wild age-progression test set.
Results for one celebrity (Tom Cruise) are shown in the main paper (Figure 2), while the remaining four—Brad Pitt, Leonardo DiCaprio, Matt Damon, and Robert Downey Jr.—are presented in Figure~\ref{supp:cylic-figure}.
Across all age cycles, our method consistently yields higher identity similarity than HRFAE, CUSP, and FADING, demonstrating stronger face aging with identity preservation in both forward and reverse age transformations.

% \begin{table}[t!]
%     \caption{Age distribution statistics of the collected celebrity dataset across predefined age groups.}
%     \label{table:celebrity-age-stats}
%     \vspace{-1em}
%     \centering
%     \begin{adjustbox}{width=0.46\textwidth}
%     \footnotesize
%     \begin{tabular}{lccc}
%         \toprule
%         Actor Name & 20--40 & 40--50 & $>$50 \\
%         \midrule
%         Brad Pitt          & 30 & 45 & 60 \\
%         Leonardo DiCaprio  & 20 & 40 & 51 \\
%         Matt Damon         & 25 & 40 & 55 \\
%         Robert Downey Jr   & 25 & 40 & 60 \\
%         Tom Cruise         & 27 & 40 & 60 \\
%         \bottomrule
%     \end{tabular}
%     \end{adjustbox}
% \end{table}

\subsection{Short-Range Aging Analysis} \label{supp:Short-Range-Aging-Analysis}
\begin{figure}[t!]
    \centering 
    \includegraphics[width=0.5\textwidth]{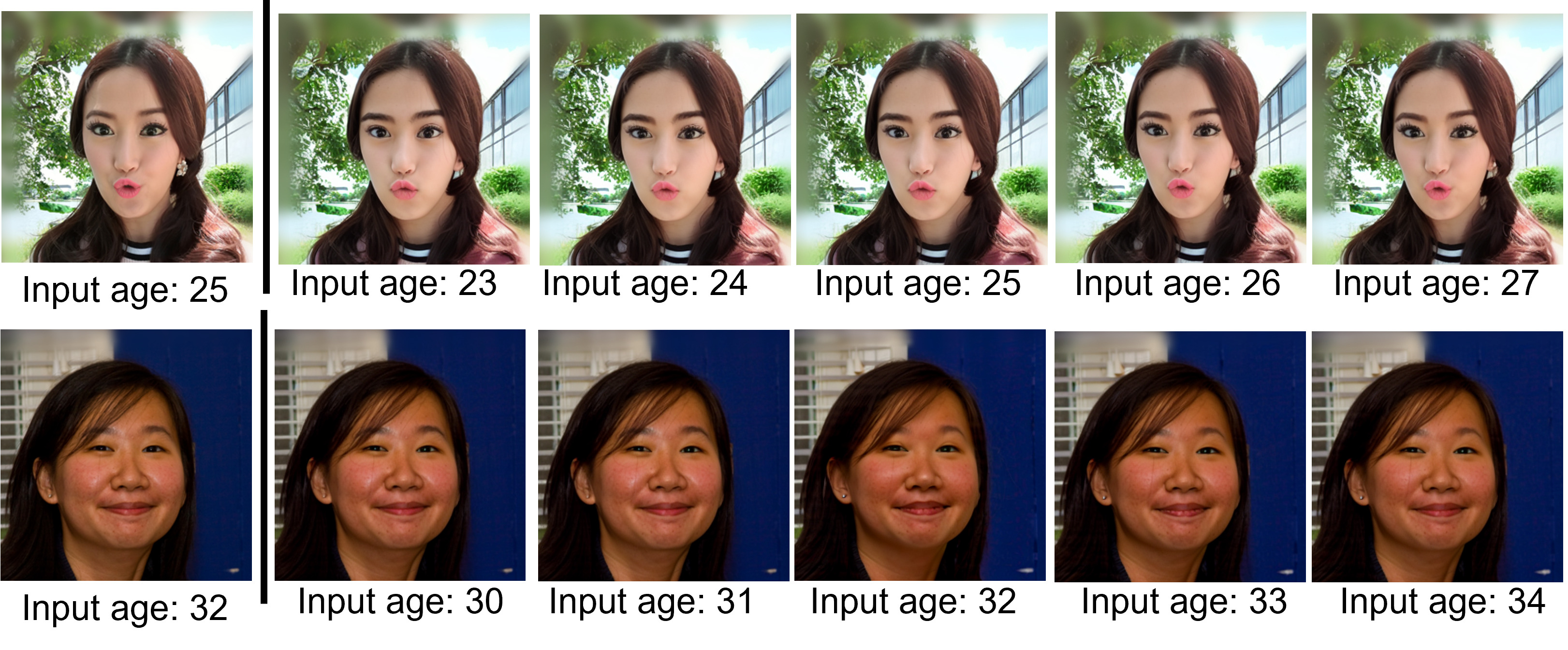}
    % \vspace{-1.8em}
    \caption{Short-range aging analysis for identity preservation. Each row shows aging results within a narrow age window (±2 years) around the input age. Our method preserves identity exceptionally well across all nearby target ages, producing nearly indistinguishable facial structure, expression, and texture from the original input. This demonstrates that even under fine-grained age perturbations—where inconsistencies are easiest to detect—FaceTT maintains stable and coherent identity features without drifting or introducing artifacts.}
    \label{fig:identity_preservation}
    \vspace{-1em}
\end{figure}

To further assess the identity-preservation capability of our method, we perform a Short-Range Aging Analysis that evaluates identity stability under fine-grained age perturbations. Figure 9 in the main paper shows examples for two identities, demonstrating consistent identity retention across closely spaced target ages. Additional results for two more identities are provided in Figure~\ref{fig:identity_preservation}, further highlighting the robustness of our approach in maintaining facial structure and appearance under small age variations.
% To further demonstrate the identity-preservation capability of our method, we conducted a Short-Range Aging Analysis to assess identity stability under fine-grained age perturbations. Examples for two identities are presented in Figure 9 of the main paper, illustrating consistent identity preservation across nearby target ages. Additional results for two more identities are provided in Figure~\ref{fig:identity_preservation}, further demonstrating the robustness of our method in maintaining facial structure and appearance under small age variations.
% we present results in Figure~\ref{fig:identity_preservation}, where an individual of age $x$ has been re-aged to nearby target ages, specifically $x-2$, $x-1$, $x$, $x+1$, $x+2$ .The results clearly illustrate that our approach maintains the individual's identity while applying realistic aging transformations. When the target age is close to the input age, our method ensures that key facial features remain unchanged, preventing unwanted distortions or identity shifts. 

\section{Additional Ablation Analysis}
\label{supp:ablation}
% Addition to ablation analysis presented in the main paper, this section presents additional ablation studies to support our contributions.
% \subsection{Analysis on FFHQ dataset}
% \label{supp:analysis-FFHQ}
% In addition to the ablation analysis presented in the main paper on the CelebHQ dataset, Table \ref{tab:abl:FFHQ} offers further analysis across different age groups on the FFHQ-Aging dataset \cite{lifespan} to validate the effectiveness of our proposed techniques. This analysis is evaluated using mean absolute error (MAE), with lower values indicating better performance.
% Regarding the inversion technique-based analysis, the proposed \textit{angular inversion} achieves a mean MAE of 11.40, outperforming all baselines, particularly excelling in the youngest (0--2 age group: 11.05) and older (50--69 age group: 11.12) age groups. For editing technique analysis, our AAC also obtains a mean MAE score of 11.40, demonstrating consistent robustness across all age groups. 

\begin{figure}[t]
    \centering
    \includegraphics[width=0.5\textwidth]{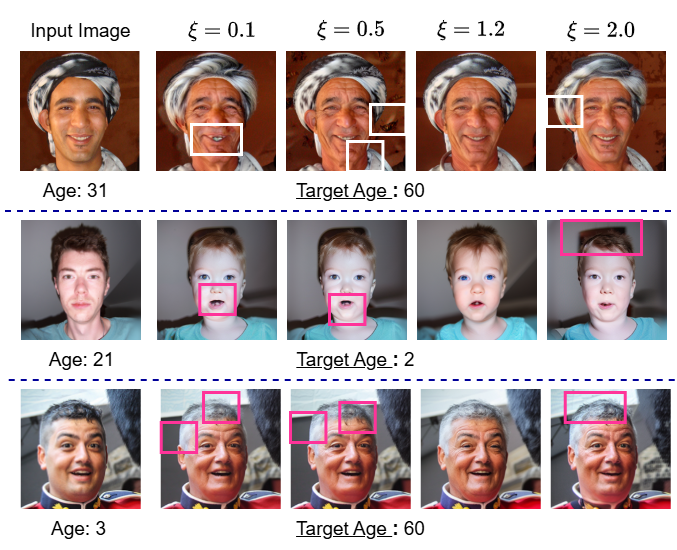}
    % \vspace{-1.5em}
    \caption{Ablation analysis on different $\xi$ values being used in the proposed \textit{Angular Inversion} technique. The results demonstrate how different $\xi$ values influence the balance between preserving identity and achieving accurate transformations to match the target prompts (e.g., age 2 or 60 years). Among the tested values, $\xi$ = 1.2 provides the best results, striking an optimal trade-off between transformation fidelity and identity preservation.
    }
    \label{fig:visual-ablation-xi}
    \vspace{-1em}
\end{figure}
\begin{figure}[t!]
    \centering 
    \includegraphics[width=0.5\textwidth]{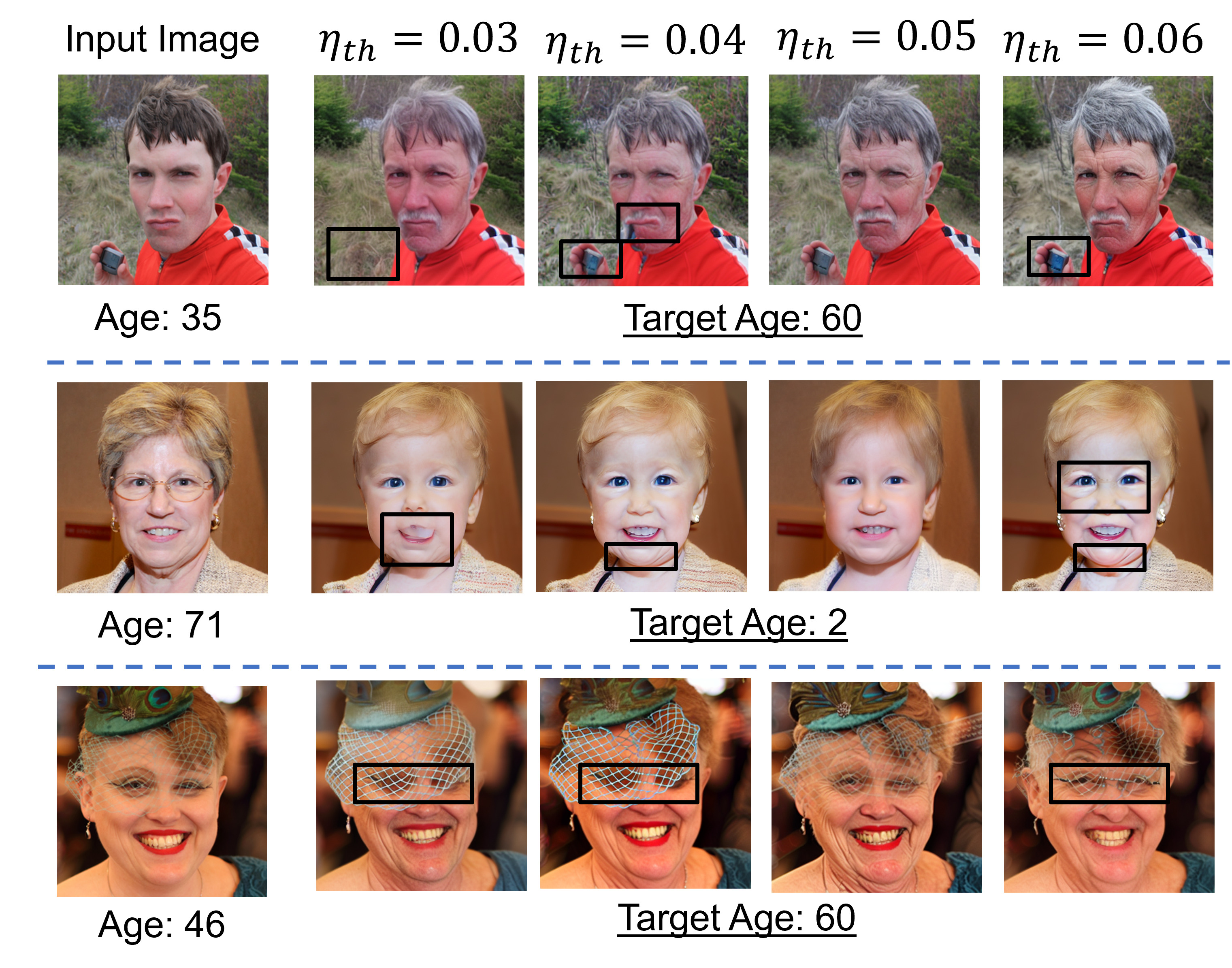}
    % \vspace{-1.8em}
    \caption{Ablation analysis on different $\eta_{th}$ values in the proposed \textit{Adaptive Attention Control} technique. The results illustrate how varying $\eta_{th}$ impacts the trade-off between maintaining structural consistency and achieving accurate aging transformations. Lower values (e.g., $\eta_{th}$ = 0.03) retain excessive source attributes, leading to incomplete aging effects, while higher values (e.g., $\eta_{th}$ = 0.06) introduce artifacts and distortions in fine details (highlighted in black boxes). The optimal value, $\eta_{th}$ = 0.05, provides the best balance, ensuring realistic aging transformations while preserving important identity and contextual details.}
    \label{fig:ablation_ita}
\end{figure}
\begin{figure}[t!]
    \centering 
    \includegraphics[width=0.47\textwidth]{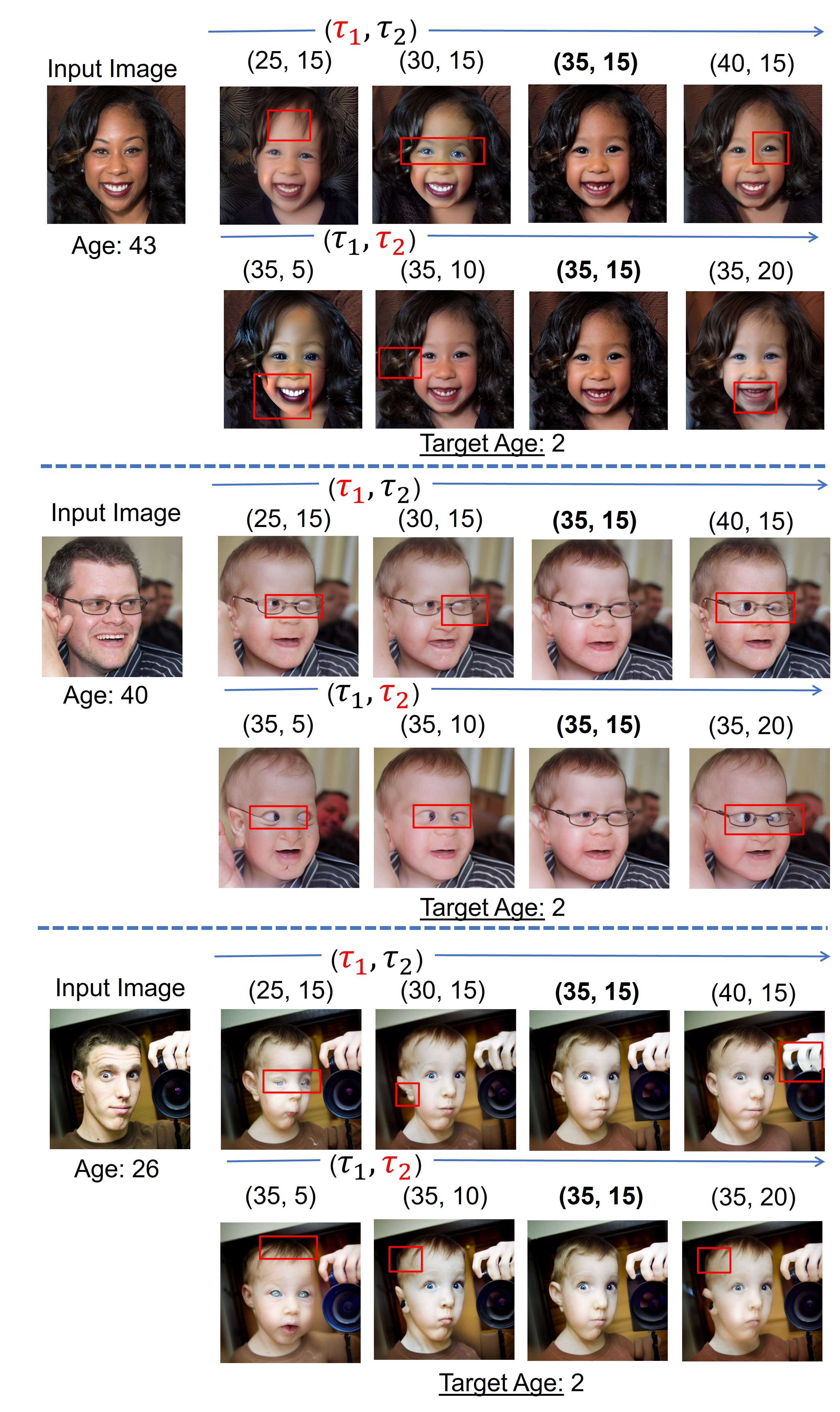}
    \vspace{-1em}
    \caption{Ablation analysis on different ($\tau_{1}$, $\tau_{2}$) values in the \textit{Adaptive Attention Control} technique, investigating their effect on balancing structural preservation and non-rigid transformations. 
    % The parameters $\tau_{1}$ and $\tau_{2}$ determine the transition between cross-attention to either cross/self-attention to self-attention adaptation during the sampling process. 
    Lower values of $\tau_{1}$ (e.g., 25) retain more source characteristics but limit the effectiveness of the transformation, while higher values (e.g., 40) can lead to excessive modifications, causing distortions in fine details (highlighted in red boxes). Similarly, an improper choice of $\tau_{2}$ can result in either overly rigid transformations or undesirable warping effects. The best balance is achieved with ($\tau_{1}$, $\tau_{2}$) = (35,15), ensuring natural age transformation while preserving key identity features.}
    \label{fig:ablation_t1_t2}
    \vspace{-1em}
\end{figure}

\begin{figure}[t!]
    \centering
    % \vspace{-0.8em}
    \includegraphics[width=0.5\textwidth]{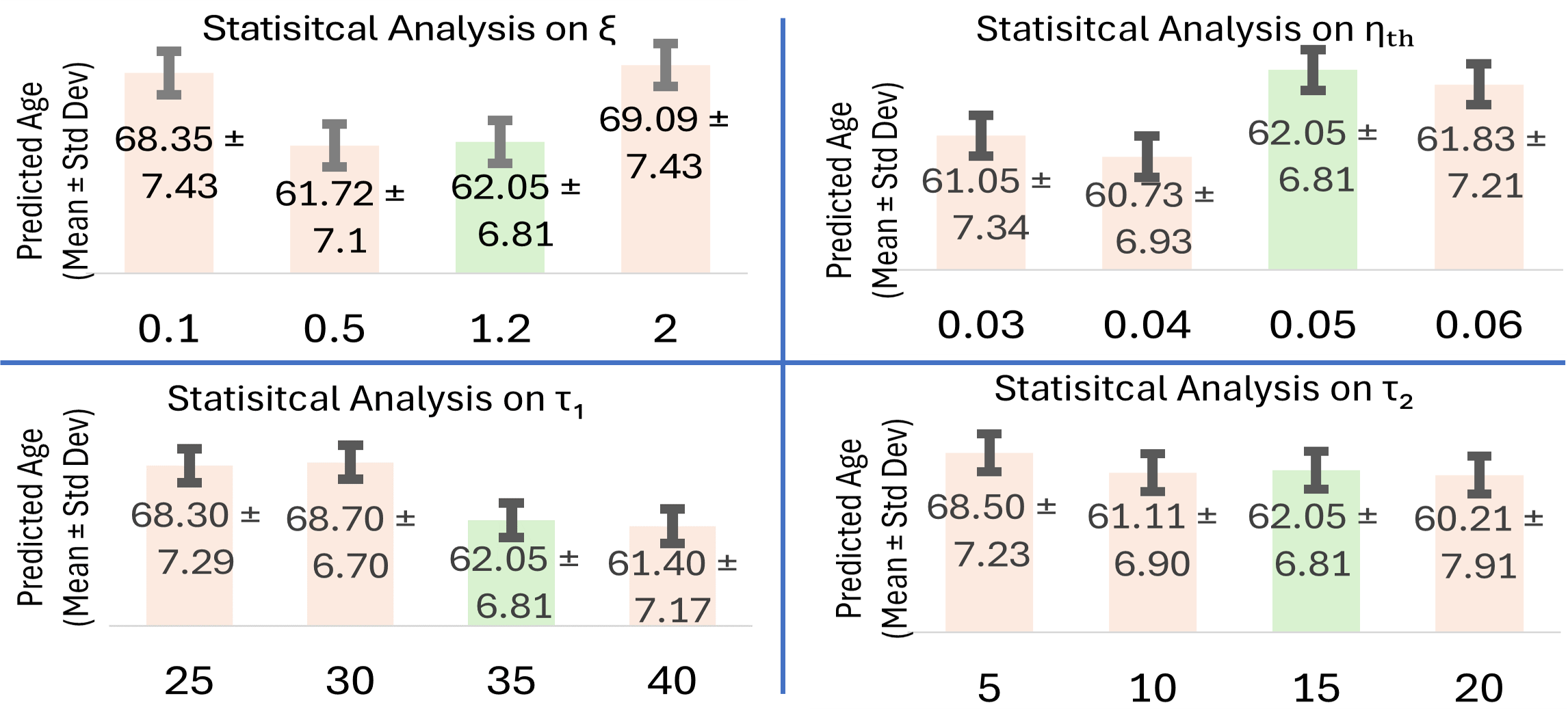}
    \vspace{-2em}
    \caption{\footnotesize Statistical analysis. Targeted predicted age: 65.14$\pm$4.86}
    \label{fig:statistical}
    % \vspace{-1em}
\end{figure}

\subsection{Visual Analysis on Hyper-Parameters}
\label{supp:visual-hyper-parameter}
Figure \ref{fig:visual-ablation-xi} analyzes the impact of the hyper-parameter \(\xi\) used in \textit{Angular Inversion}. Low \(\xi\) values (specifically, \(\xi = 0.1, 0.5\)) result in minimal changes, allowing the input identity to be preserved but producing weaker re-aging effects. Conversely, high \(\xi\) values (such as \(\xi = 2.0\)) lead to excessive alterations, causing distortions and a loss of identity. The optimal balance is found at \(\xi = 1.2\), which provides realistic re-aging while maintaining identity, as shown in the first and third rows. This highlights \(\xi = 1.2\) as the preferred choice for high-quality results.

Figure \ref{fig:ablation_ita} examines the effect of $\eta_{th}$ in \textit{Adaptive Attention Control}, which determines the transition between self-attention and cross-attention adaptation. When $\eta_{th}$ is too low (e.g., 0.03), the attention mechanism primarily preserves source characteristics, leading to insufficient aging effects. Conversely, setting $\eta_{th}$ too high (e.g., 0.06) results in excessive modification of details, introducing artifacts. The best balance occurs at $\eta_{th}$ = 0.05, enabling natural aging transformations while maintaining identity consistency.

Figure \ref{fig:ablation_t1_t2} evaluates the impact of ($\tau_{1}$, $\tau_{2}$), which regulates the selection of either cross-attention or self-attention or mixture of both (based on the value of $\eta_{th}$) in \textit{Adaptive Attention Control}. Low $\tau_{1}$ values (e.g., 25) retain too many source attributes, preventing effective aging transformations. On the other hand, high values (e.g., 40) lead to exaggerated changes, distorting facial features. Similarly, improper selection of $\tau_{2}$ affects the smoothness of the transformation. The optimal configuration, ($\tau_{1}$, $\tau_{2}$) = (35,15), provides the best trade-off between structural consistency and realistic aging effects, as demonstrated in Figure \ref{fig:ablation_t1_t2}.

% These analyses collectively highlight the importance of tuning these hyper-parameters to achieve high-fidelity, identity-preserving age transformations.

\subsection{Statistical Analysis of Hyper-Parameter}
\label{supp:statistical-hyper-parameter}
To further validate our approach, we incorporated a statistical analysis in Fig.~\ref{fig:statistical}, where the mean ± standard deviation of the predicted ages is reported for different age groups across various hyperparameter configurations. This evaluation was conducted on the CelebA-HQ dataset to assess the consistency and accuracy of the generated age transformations. The statistical trends observed support the robustness of our chosen hyperparameters, ensuring realistic and demographically appropriate age progression across the dataset.

% \begin{figure}[t!]
%     \centering 
%     \includegraphics[width=0.475\textwidth]{Images/Face-Reaging-cycle-identity.png}
%     \vspace{-1.5em}
%     \caption{Cycle-consistency results across four individuals demonstrate strong identity preservation under extreme age transformations. Top: Faces initially old are de-aged and then re-aged back, closely matching the original input. Bottom: Young faces are re-aged and subsequently de-aged, successfully recovering the original appearance. Key identity traits—such as facial structure, eye shape, and overall geometry—remain visually consistent, indicating minimal drift despite large age gaps.}
%     \label{fig:cycle-identity_preservation}
%     \vspace{-1em}
% \end{figure}

\section{Analysis on Inversion \& Editing Techniques}
\label{supp:visual-inversion-editing}
\begin{figure*}[t]
    \centering
    % \hspace{-0.5cm}
    \includegraphics[width=0.84\textwidth]
    {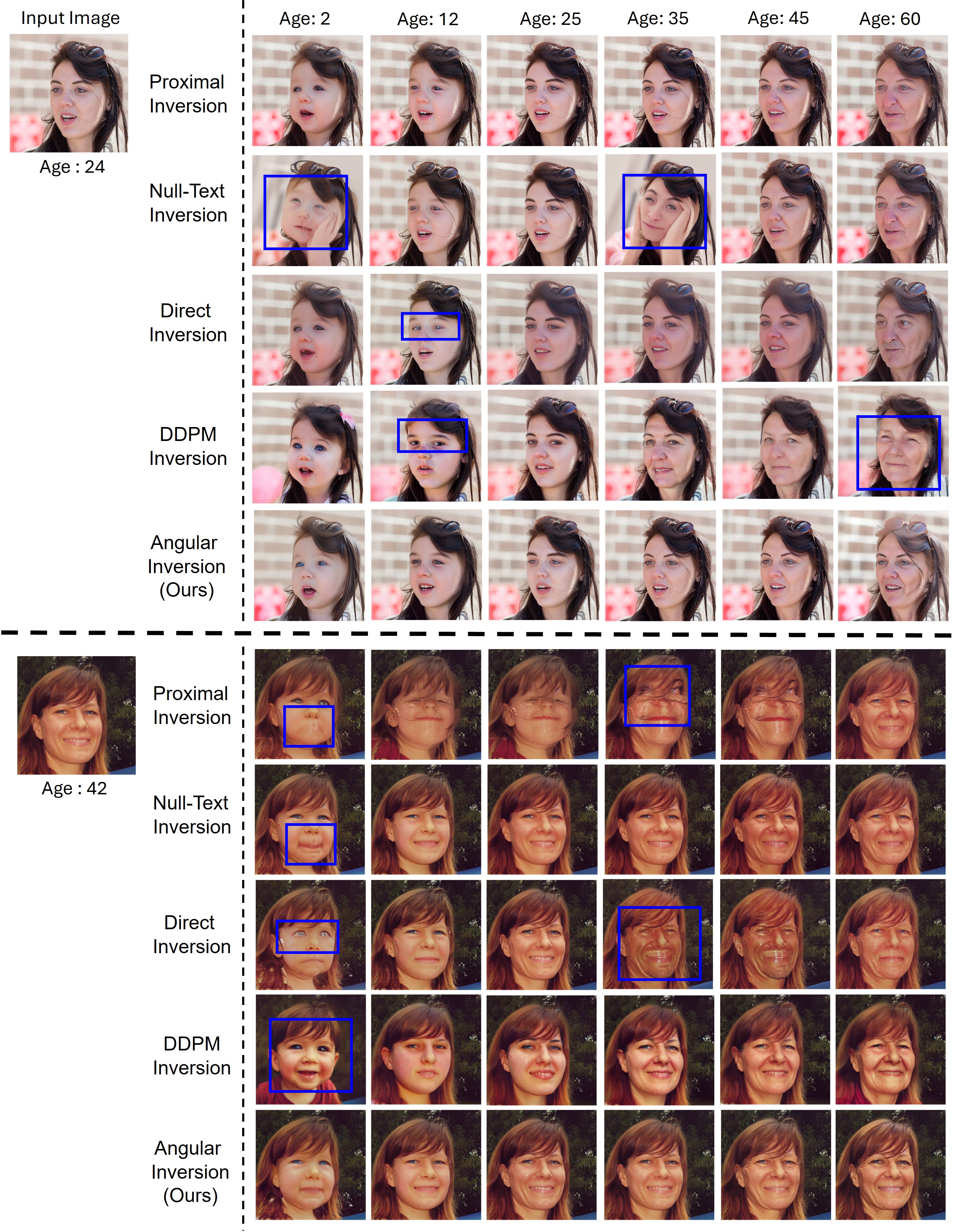}
    % {images/Supp_Ablation1_Inversion_Visual_updated.pdf}
    \vspace{-1em}
    \caption{Comparison of different inversion methods for facial re-aging across multiple target ages (2 → 60 years). Proximal, Null-Text, Direct, and DDPM inversion each introduce distortions or identity drift during the re-aging process. Proximal Inversion often produces inconsistent geometry and blurry structural changes, leading to unnatural aging transitions. Null-Text Inversion exhibits significant shape warping and texture artifacts \textcolor{blue}{(blue boxes)}, particularly in early-age predictions. Direct Inversion suffers from identity inconsistency and incorrect facial proportions across ages, while DDPM Inversion frequently alters key identity features \textcolor{blue}{(blue boxes)}, causing the re-aged outputs to deviate from the input subject. In contrast, \textit{Angular Inversion} yields stable, high-fidelity reconstructions with smooth and realistic age progression, maintaining consistent identity cues from childhood to old age. The results highlight that \textit{Angular Inversion} achieves the most reliable and artifact-free initialization for age editing, enabling accurate transformations across the entire age spectrum.}
    \label{fig:visual-ablation-inversion-supp1}
    \vspace{-1em}
\end{figure*}

The additional analysis is illustrated in Figures~\ref{fig:visual-ablation-inversion-supp1}, which shows a visual comparison between the proposed \textit{Angular Inversion} and existing inversion techniques \cite{proximal, negative-prompt, null-text, direct-inversion}. This comparison shows that the Proximal Inversion \cite{proximal} achieves partial alignment with target prompts but struggles with fine details, resulting in inconsistent transformations. While Negative-Prompt Inversion \cite{negative-prompt} enhances alignment, it introduces distortions, particularly in extreme age changes. Null-Text inversion \cite{null-text} provides better prompt alignment but compromises identity and realism, leading to overly smoothed features. Direct Inversion \cite{direct-inversion} maintains facial structure but lacks the adaptability needed for extreme transformations, which results in artifacts. In contrast, our \textit{Angular Inversion} method captures fine details and preserves semantic alignment and identity, resulting in realistic and coherent age transformations. Its balanced performance proves its superiority over existing methods.

\begin{table}[t]
\centering
\caption{Comparative analysis on CelebA-HQ dataset for young-to-60 task. Best score: in \colorbox{red!20}{red}, second-best score: in \colorbox{celadon!50}{green}.} \label{tab:table6}
\vspace{-1em}
\begin{adjustbox}{width=0.49\textwidth}
\begin{tabular}{lcccccc}
\toprule
\textbf{Techniques} & \textbf{Predicted age} & \textbf{Blur $\downarrow$} & \textbf{Gender $\uparrow$} & \textbf{Smiling $\uparrow$} & \textbf{Neutral $\uparrow$} & \textbf{Happy $\uparrow$} \\ \midrule
\multicolumn{7}{l}{Analysis on Inversion techniques} \\ \midrule
Real Images & 65.14 $\pm$ 4.86 & 1.73 & --- & --- & --- & --- \\
Null-Text~\cite{null-text} & 69.88 $\pm$ 6.20 & 2.18 & 98.44 & \colorbox{celadon!50}{76.17} & 56.54 & \colorbox{red!20}{73.19} \\
Negative~\cite{negative-prompt}& 69.35 $\pm$ 5.51 & 2.69 & 98.23 & 74.35 & 61.77 & 69.19 \\
Proximal~\cite{proximal} & 68.39 $\pm$ 6.74 & \colorbox{celadon!50}{2.49 }& 98.13 & 73.99 & \colorbox{celadon!50}{63.94} & 71.98 \\
Direct~\cite{direct-inversion} & \colorbox{celadon!50}{61.84 $\pm$ 6.80} & 2.18 & \colorbox{celadon!50}{99.71} & 73.74 & \colorbox{red!20}{64.65} & 66.23 \\
% \rowcolor{lightmintbg} 
\textit{Angular} (Proposed) & \colorbox{red!20}{62.05 $\pm$ 6.81} & \colorbox{red!20}{2.18} & \colorbox{red!20}{99.79} & \colorbox{red!20}{78.31} & 62.67 & \colorbox{celadon!50}{72.17}  \\ \midrule
\multicolumn{7}{l}{Analysis on editing techniques} \\ \midrule
Real Images & 65.14 $\pm$ 4.86 & 1.73 & --- & --- & --- & --- \\
P2P~\cite{p2p}& 57.31 $\pm$ 4.23 & 2.31 &  97.40 & 72.07 & 61.19 & 65.77  \\
MasaCtrl~\cite{masactrl} & 57.23 $\pm$ 7.65 & 2.57 &  96.90 & 75.39 & \colorbox{red!20}{64.25} & 66.32  \\
PnP~\cite{pnp} & 69.35 $\pm$ 6.26  & 2.30 & \colorbox{celadon!50}{98.67} & \colorbox{celadon!50}{78.22} & 62.15 & \colorbox{celadon!50}{69.85}  \\
FPE~\cite{FPE} & \colorbox{celadon!50}{68.45 $\pm$ 6.32 }&\colorbox{celadon!50}{ 2.27} &  97.53 & 73.39 & 62.36 & 67.20  \\
% \rowcolor{lightmintbg} 
\textit{AAC} (Proposed) & \colorbox{red!20}{62.05 $\pm$ 6.81} & \colorbox{red!20}{2.18} & \colorbox{red!20}{99.79} & \colorbox{red!20}{78.31} & \colorbox{celadon!50}{62.67} & \colorbox{red!20}{72.17}  \\ \bottomrule
\end{tabular}
\end{adjustbox}
\vspace{-1.2em}
\end{table}

\begin{figure*}[t]
    \centering
    % \hspace{-0.6cm}
    \includegraphics[width=0.84\textwidth]{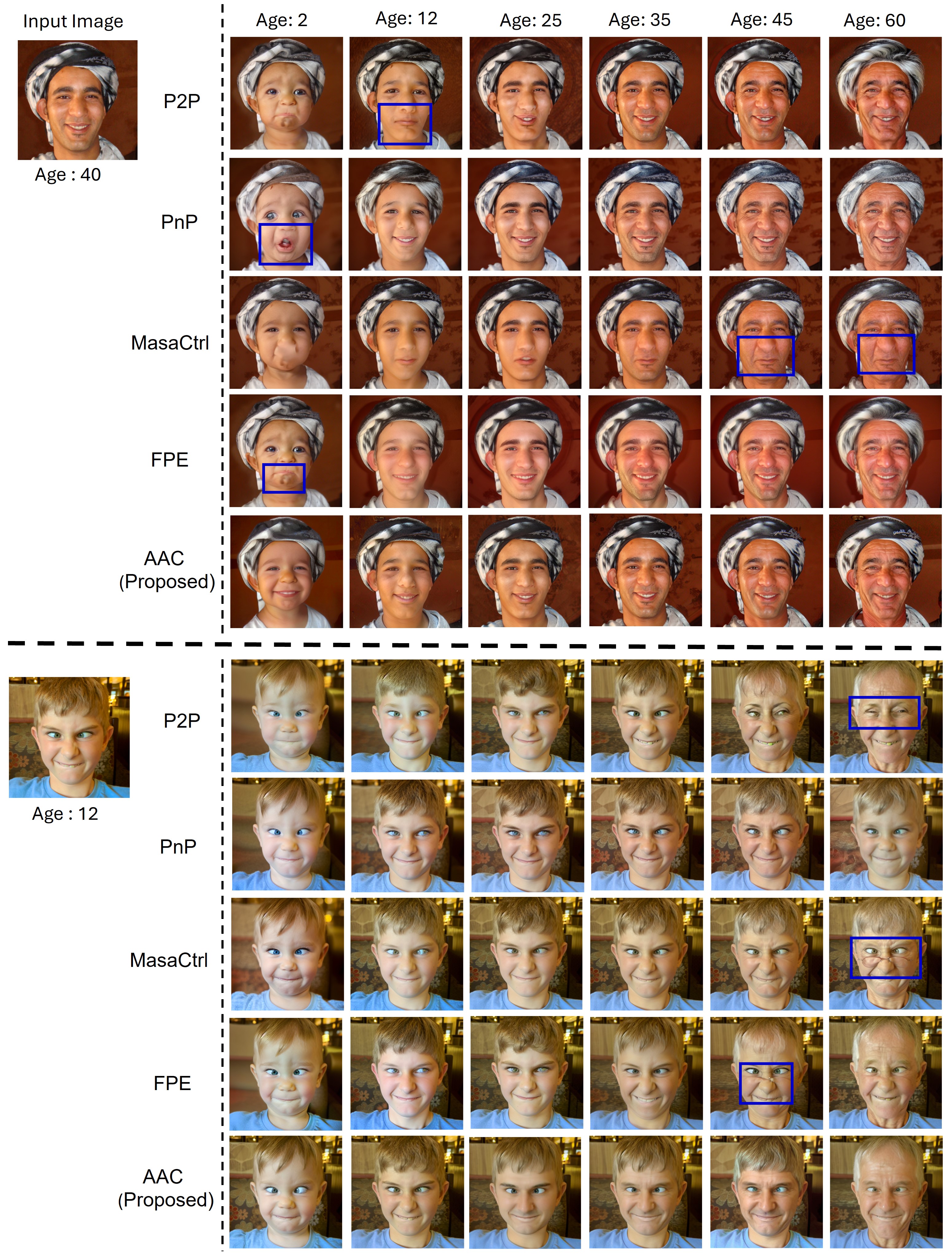}
    \vspace{-1em}
    \caption{Comparison of age editing using P2P, PnP, MasaCtrl, FPE, and our \textit{Adaptive Attention Control (\textbf{AAC})} across target ages 2→60. Existing attention-based editing methods exhibit noticeable artifacts and identity inconsistencies: P2P often produces over-smoothed faces and distorted geometry \textcolor{blue}{(blue boxes)}, especially for younger ages; PnP struggles with preserving identity and frequently introduces incorrect mouth and eye structures; MasaCtrl shows unstable attention modulation, leading to drifted identity and inconsistent aging patterns; and FPE tends to hallucinate textures or distort facial regions \textcolor{blue}{(blue boxes)}, particularly in older targets. In contrast, \textit{AAC} achieves smooth and coherent age progression while consistently preserving identity, facial structure, and attribute integrity across the entire age spectrum. The results demonstrate that Adaptive Attention Control provides more precise, stable, and semantically aligned guidance than prior attention-editing approaches.}
    \label{fig:visual-ablation-editing2}
    \vspace{-1em}
\end{figure*}

Figures~\ref{fig:visual-ablation-editing2} show an additional comparison of the proposed \textit{AAC} with existing image editing techniques, including P2P \cite{p2p}, PnP \cite{pnp}, MasaCtrl \cite{masactrl}, and FPE \cite{FPE}. It shows that P2P struggles to balance alignment and identity, often resulting in exaggerated or distorted features. While PnP is more consistent, it is still prone to artifacts, especially in extreme cases. MasaCtrl improves prompt fidelity but tends to overemphasize features, resulting in unnatural transformations. FPE balances structure and identity but lacks adaptability in extreme cases, resulting in overly smooth results. In contrast, our \textit{AAC} outperforms these methods by modulating attention mechanisms to achieve semantic fidelity and identity preservation. It delivers realistic transformations, handling challenging cases like extreme age regression or progression with natural, coherent outputs.

\section{Ethical Concerns} \label{supp:ethical}
The development of facial re-aging technology raises several ethical considerations that must be addressed to ensure responsible usage. The ability to manipulate facial features across different age groups brings up concerns about privacy, identity misuse, and potential exploitation. For example, re-aging technology could be misused to create deceptive content, such as deepfakes, which might lead to misinformation or harm an individual's reputation. This challenge is applied to all image editing methods in general. However, advancements in detecting and mitigating malicious edits are evolving quickly. We believe our work will support these efforts by providing insights and access of the proposed image editing and generation process.

\section{Baseline Details} \label{supp:baselines}
Our proposed framework for facial re-aging is compared with three SOTA methods \cite{hrfae,cusp,FADING}. Details of these baselines are as follows:
\begin{itemize}
    \item \textbf{HRFAE \cite{hrfae}}, a hybrid model that uses feature-aligned encoders to preserve identity and achieve photorealistic re-aging effects. However, HRFAE often struggles with extreme age transformations, leading to minor inconsistencies in older age predictions.
    \item \textbf{CUSP \cite{cusp}} combines cycle-consistent adversarial networks with spatial priors to achieve smooth age progression and regression. Despite its ability, CUSP sometimes fails to capture fine-grained aging details accurately. 
    \item \textbf{FADING \cite{FADING}} uses null-text inversion \cite{null-text} and attention control for facial image editing with a pre-trained diffusion model. Despite high-quality results, it faces challenges with identity consistency in extreme age edits and is computationally intensive for real-time applications.
\end{itemize}

\noindent To validate the effectiveness of our proposed \textit{angular inversion} technique, we compare it with existing inversion techniques \cite{null-text,negative-prompt,proximal,direct-inversion}.

\begin{itemize}
    \item \textbf{Null-Text Inversion \cite{null-text}} uses pivot tuning with null-text prompts for better alignment but suffers from inefficiencies in computation. 
    \item \textbf{Negative-Prompt Inversion \cite{negative-prompt}} speeds up inversion by approximating DDIM inversion, but sacrifices reconstruction quality. 
    \item \textbf{Proximal Inversion \cite{proximal}} improves Negative-Prompt Inversion by using proximal guidance, but struggles with latent space disruptions, particularly for complex transformations.
    \item \textbf{Direct Inversion \cite{direct-inversion}} separates source and target branches for better content preservation but faces challenges with complex edits, especially when large attribute shifts are required.
\end{itemize}

\noindent We also validate the proposed \textit{AAC} editing technique with existing baselines \cite{p2p,pnp,masactrl,FPE}.
\begin{itemize}
    \item \textbf{Prompt-to-Prompt (P2P) \cite{p2p}} adjusts cross-attention to maintain spatial consistency in edits. Although effective in many scenarios, it struggles with transformations requiring significant structural changes.
    \item \textbf{Plug-and-Play (PnP) \cite{pnp}} modifies attention maps for text-driven image editing but introduces minor artifacts. 
    \item \textbf{MasaCtrl \cite{masactrl}} enables mutual self-attention control but generates noticeable artifacts in challenging transformations. 
    \item \textbf{FPE \cite{FPE}} modifies self-attention maps for stable edits but struggles with precise age-specific effects. 
\end{itemize}